\documentclass{article}

\usepackage{PRIMEarxiv}

\usepackage[utf8]{inputenc} 
\usepackage[T1]{fontenc}    
\usepackage{url}            
\usepackage{booktabs}       
\usepackage{amsfonts}       
\usepackage{nicefrac}       
\usepackage{microtype}      
\usepackage{lipsum}
\usepackage{fancyhdr}       
\usepackage{graphicx}       

\newcommand{\methodsref}[1]{\hyperref[#1]{Methods}}

\usepackage[title]{appendix}
\usepackage{amsmath}
\usepackage{amsfonts}
\usepackage{amssymb}
\usepackage{graphicx}
\usepackage{siunitx}
\usepackage{xcolor}
\usepackage{booktabs}
\usepackage{caption}
\usepackage{float}
\usepackage{color,soul}
\usepackage{hyperref}
\hypersetup{
    colorlinks=true,
    linkcolor=blue,
    filecolor=blue,      
    urlcolor=blue,
    pdftitle={Overleaf Example},
    pdfpagemode=FullScreen,
    }

\usepackage{bm}

\usepackage{comment}

\title{Equilibrium Conserving Neural Operators for Super-Resolution Learning}

\author{
  Vivek Oommen$^{1,+}$, Andreas E. Robertson$^{2,+}$, Daniel Diaz$^3$, Coleman Alleman$^{4}$, Zhen Zhang$^5$,\\ \textbf{Anthony D. Rollett}$^3$, \textbf{George E. Karniadakis}$^5$, \textbf{R\'emi Dingreville}$^{2,}$\thanks{Corresponding author: rdingre@sandia.gov}  \\ \\
  1-School of Engineering, Brown University, Providence, RI 02912, USA \\
  2-Center for Integrated Nanotechnologies, Sandia National Laboratories, Albuquerque, NM 87185, USA \\
  3-Department of Materials Science and Engineering, Carnegie Mellon University, Pittsburgh, PA 15123, USA \\
  4-Mechanics of Materials Department, Sandia National Laboratories, Livermore, CA 94550, USA \\
  5-Division of Applied Mathematics, Brown University, Providence, RI 02912, USA \\ \\
  +-these authors contributed equally to this work, order selected randomly. \\
}

\begin{document}
\maketitle

\begin{abstract}
Neural surrogate solvers can estimate solutions to partial differential equations in physical problems more efficiently than standard numerical methods, but require extensive high-resolution training data. In this paper, we break this limitation; we introduce a framework for super-resolution learning in solid mechanics problems.  Our approach allows one to train a high-resolution neural network using only low-resolution data. Our Equilibrium Conserving Operator (ECO) architecture embeds known physics directly into the network to make up for missing high-resolution information during training. We evaluate this ECO-based super-resolution framework that strongly enforces conservation-laws in the predicted solutions on two working examples: embedded pores in a homogenized matrix and randomly textured polycrystalline materials. ECO eliminates the reliance on high-fidelity data and reduces the upfront cost of data collection by two orders of magnitude, offering a robust pathway for resource-efficient surrogate modeling in materials modeling. ECO is readily generalizable to other physics-based problems.
\end{abstract}

\keywords{Physics-Informed Machine Learning $|$ Hard Constraining Physics $|$ Super-resolution $|$ Elasticity}

\section*{Introduction}
\label{sec: intro}
{
Physics-based numerical simulations allow scientists to systematically and thoroughly analyze complex multiscale physical processes that dictate material behavior, from atomic interactions to macroscale engineering phenomena.
Crystal-plasticity simulations \cite{roters2010cpfemoverview}, for instance, have proven invaluable in manufacturing where by linking processing parameters to microstructural features they were used to identify the root causes of earring defects in deep drawing processes \cite{raabe2005_earingcpfem}.
However, the computational demands of such simulations present significant challenges.
Traditional numerical schemes, especially three-dimensional (3D) finite element methods, face crippling computational demands due to the exponential scaling of their degrees of freedom with respect to spatial dimensions \cite{hughes2003finite}.
This scaling behavior drives reliance on
supercomputing clusters \cite{NASA_HEC_2020} with massive energy footprints \cite{lannelongue2021greenalgorithms},
advanced parallelization techniques \cite{garcia2019mpi}, and
heterogeneous hardware solutions \cite{lei2023heterogeneous}.
Yet, even with these workarounds, applications of simulations are still restricted to low resolutions and small domains \cite{raabe2004continuum, mcdowell2018microstructure}, particularly in applications requiring many repeated solves, e.g., microstructure-sensitive design \cite{fullwood2010microstructure}.
}

{
Neural-network-based surrogate solvers have emerged as an alternative approach for efficiently solving complex partial differential equations (PDEs).
Acceleration is achieved because the neural solver displays orders of magnitude lower inference costs than direct numerical simulation (DNS) \cite{azizzadenesheli2024neural}.
Several neural surrogate strategies are being actively researched, including physics-informed neural networks \cite{raissi2019physics},
neural operators \cite{li2020fourier, lu2021learning, cao2023lno, tripura2023wavelet}, and 
vectorized neural networks \cite{gupta2022towards, ovadia2023ditto}.
At the time of this writing, no single dominant methodology has been established; rather, ongoing research efforts focus on developing best practices to address certain important characteristics, including parametric PDEs and training data cost.
}

{
One significant area of exploration -- particularly in neural surrogates for computational mechanics -- involves managing parametric PDEs under nonstationary input conditions, such as fluctuating boundary constraints and spatially heterogeneous microstructure fields.
These parameters define parametric PDE families.
Operator learning \cite{kelly2024therino, you2022learning} and vectorized neural network methods \cite{kelly2021recurrent, brady2024unet} learn conditional mappings from these high-dimensional inputs (e.g., microstructural descriptors) to PDE solution fields.
A notable challenge in surrogate modeling for solid mechanics is that predictive accuracy hinges on resolving the parametric dependence on material microstructures \cite{ghosh2011computational}.
To address this complexity and achieve model support over this high-dimensional parametric space, neural network-based surrogate solvers often rely on data-driven strategies necessitating large training datasets of input-output pairs derived from traditional numerical solvers \cite{khorrami2023unet}.
The significant cost of curating such large training datasets often practically offsets any downstream gains due to faster solving speeds.
}

{
Physics-informed learning is a promising strategy for addressing the data-intensive requirements of traditional neural networks by supplementing reduced training data with domain-specific physical principles \cite{wang2021learning, li2024physics, goswami2023physics}.
Popularized by Physics-Informed Neural Networks (PINNs) \cite{toscano2024pinns, penwarden2022multifidelity}, traditionally, physics-informed machine learning methods minimize the surrogate prediction's deviations from the governing physics by weakly penalizing the PDE residual in the loss function.
Although PINNs -- used to solve single PDEs -- rely only on the PDE residual for training, physics-informed extensions to continuous \cite{goswami2023physics, li2024physics} and vectorized \cite{medvedev2025physics} operator methods for solving parametric PDEs rely on a mixture of the PDE residual and standard supervised training data. 
By penalizing deviations from governing PDEs \cite{raissi2019physics, goswami2023physics, li2024physics}, these approaches reduce reliance on large training data requirements \cite{wang2021learning, li2024physics} while simultaneously improving physical plausibility of the solution for downstream tasks \cite{toscano2024pinns} and improving inference performance in regions with limited supervised data \cite{li2024physics, toscano2024pinns}.
Although theoretically simple, the PINN approach can be challenging to implement due to poor training dynamics caused by the large spectral radii of common PDEs \cite{wang2021eigenvector, boudec2024pinnsarestiff}.
Recent advances have shown improved performance by directly embedding the physics into the network architecture using strong constraints \cite{lu2021physics}.
For instance, Lu \textit{et al.} demonstrated improved roll-out by using a neural network to perturb forward Euler updates in a time-dependent system \cite{lu2018beyond}.
Incorporating generalized physics principles can also be beneficial when system-specific physics are uncertain or when developing strong constraints is theoretically complex.
For example, Hamiltonian neural networks and GENERIC formalism informed neural networks (GFINNs) incorporate energy conservation and strictly increasing entropy into neural network methods for dynamic systems \cite{greydanus2019hamiltonian, lee2021structurepreserving, lee2022structurepreserving, zhang2022gfinns, zhang2022structure}.
}

{
Multi-resolution learning is another solution aimed at reducing the upfront cost of collecting training data \cite{li2020fourier}.
Unlike vectorized methods which operate on a consistent discretized space (i.e., inference occurs on the same grid resolution as training), neural-operator methods are `mesh independent'; they learn mappings between generalized function spaces from the same discrete data \cite{raonic2023convolutionalneuraloperator, li2020fourier, lu2021learning, alkin2024universal}.
Practically, this means that neural-operator-based networks can be trained on low resolution (i.e., coarsely sampled) data and used in inference for high resolution queries.
Because of the exponential scaling of DNS with increasing resolution, the ability to train on low-resolution data can drastically reduce the upfront cost of collecting data.
However, current neural-operator solutions have an important limitation.
They assume that even the coarsely sampled training data is sufficiently sampled to capture the highest frequencies present in the high-resolution representation \cite{raonic2023convolutionalneuraloperator}.
In other words, they assume the training data is sampled on a coarse grid whose Nyquist frequency is still greater than the intrinsic frequency of the underlying physical system of interest. 
Operators trained on data sampled below this intrinsic frequency limit predict oversmoothed PDE predictions \cite{harandi2024FNO_somesuperresolution, oommen2024integrating}.
The intrinsic frequency limit of some commonly studied systems -- such as low Reynold's number laminar flows \cite{li2020fourier} -- is quite low, meaning that training data resolutions can be pushed down to significantly reduce the cost of data collection.
Unfortunately, solid mechanics problems do not fall into this category.
The presence of sharp discontinuities in their input parameters (e.g., grain boundaries) leads to rich high-frequency spectra and high intrinsic frequency limits, requiring fine mesh with high Nyquist frequency for proper representation.
Here, missing high frequencies are important features such as strain energy concentrations that seed nonlinear deformation \cite{mcdowell_strainenergy} for instance.
As we will see in this paper, very limited gains are achievable using the standard operator methods.
}

{
In this paper, we introduce a framework for super-resolution learning in solid mechanics aimed at addressing resolution limitations in microstructure simulations.
Our approach leverages physics to break the Nyquist limit of low-resolution grids, enabling high-resolution neural surrogate solvers to be trained exclusively on low-resolution supervised data, thereby significantly lowering the computational costs of obtaining high-resolution training data.
A key attribute of this approach is the ability to utilize low-resolution training data sampled below the high-resolution ground truth's Nyquist frequency, which is particularly important for simulations containing high-frequency features.
We propose to compensate for the missing high-frequency data by incorporating the system's physics through physics-informed machine learning techniques.
Specifically, we formulate super-resolution learning as a physics-informed learning problem: high-resolution network predictions of kinematic and kinetic fields based on high-resolution input microstructures are penalized in two ways.
First, a physics-informed scheme at the natural high resolution and second, after coarsening the prediction, a traditional supervised scheme using the low-resolution training data. 
For the physics-informed learning element, we propose an Equilibrium Conserving Operator (ECO): a conservation law prioritizing implementation.
ECO incorporates the conservation laws (i.e., the PDEs) strongly through novel network architecture blocks that strictly enforce stress equilibrium and deformation compatibility.
In exchange, the constitutive coupling between the kinematic and kinetic fields is softly penalized in the loss function. 
The ECO formulation shifts the physics-informed machine learning paradigm from traditional soft PDE penalization (as in PINNs) to hard PDE constraints (mechanical equilibrium/deformation compatibility), while focusing the loss function on the microstructure itself and the resulting constitutive law -- the only source of high-frequency supervision -- to enable high-resolution training.
Our method offers a promising solution to the challenge of exponentially slower high-resolution simulations, as low-resolution simulations are significantly faster to generate.
By lowering the lower limit on low-resolution data requirements, we effectively eliminate the substantial upfront training cost that has traditionally plagued neural network methods in this domain.
We contrast the performance of our physics-based neural super-resolution framework implemented using either constitutive-law (traditional) or conservation-law (ECO) enforced physics-informed learning on two working examples: embedded pores in a homogenized matrix and randomly textured polycrystalline materials.
In addition, we contrast the framework's performance against a traditional neural operator-based solution.
}

\begin{figure*}[h!]
  \centering
  \includegraphics[width=0.9\textwidth]{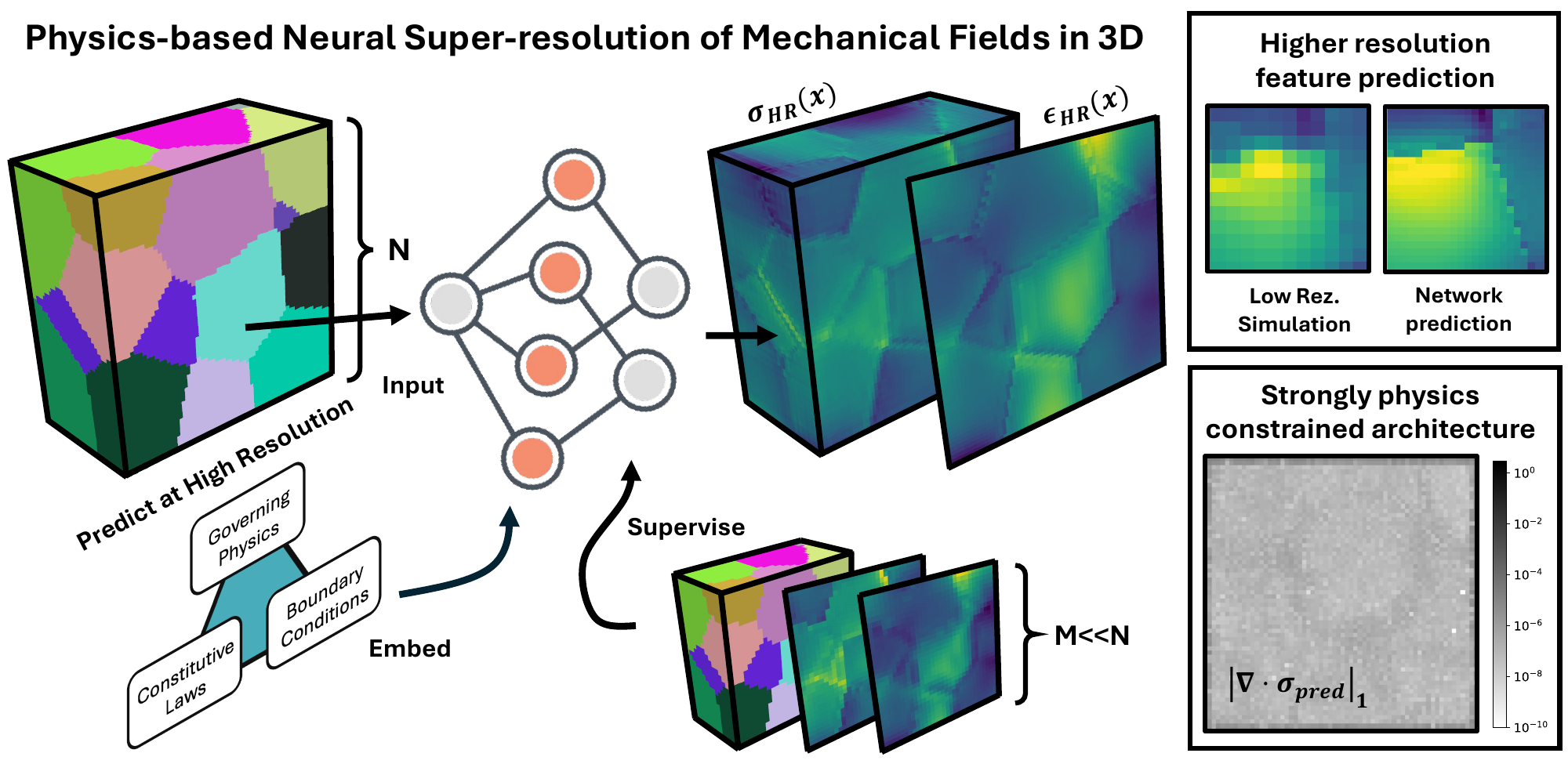}
  \caption{Overview of the physics-based neural super-resolution framework.}
  \label{fig:framework_overview}
\end{figure*}

\section*{Results}
\label{sec: results}
{
At a high level, we aim to develop a neural surrogate model that predicts the mechanical response of a material microstructure under some boundary conditions (uniaxial deformation in this case).
The model takes the microstructure topology ($\boldsymbol{m}(x)$) as input and predicts the values of kinetic variables (stresses denoted as $\sigma_{ij}$) and kinematic variables (elastic strains denoted as $\epsilon_{ij}$ or elastic deformation gradients denoted as $F^e_{ij}$) across the entire microstructure.
Acquiring large, high-resolution micromechanics datasets for training these surrogates is prohibitively expensive due to the poor scalability of traditional DNS, especially in 3D.
Consequently, we assume here that only low-resolution training data is available.
Setting aside the underlying physics, we formulate a data-driven learning problem such that:
}
\begin{equation}
    f_\theta : \boldsymbol{m} \in\mathbb{R}^{R\times N^3} \mapsto \boldsymbol{\sigma} \in \mathbb{R}^{3\times3\times N^3}, \boldsymbol{\epsilon} \in \mathbb{R}^{3\times3\times N^3},~
    \textrm{s.t.} \ \theta = \arg \min_{\theta} \mathbb{E}_{\boldsymbol{m}, \boldsymbol{\sigma}_{\rm{LR}},\boldsymbol{\epsilon}_{\rm{LR}}} \left[ L(D(f_\theta (\boldsymbol{m})), \{ \boldsymbol{\sigma}_{\rm{LR}}, \boldsymbol{\epsilon}_{\rm{LR}} \}) \right].\label{eq:mathform}
\end{equation}
{
\noindent Here, the continuous domain has been discretely sampled and represented as a high-resolution $N^3$ 3D voxel grid.
$R$ is the dimensionality of the microstructure representation at each voxel (e.g., a pore microstructure is represented by a single indicator function, therefore $H=1$. See \autoref{app:polycrystalline_inputs} for the optimal definition of $R$ for polycrystalline systems).
$\boldsymbol{\sigma}_{\rm{LR}}$ and $\boldsymbol{\epsilon}_{\rm{LR}}$ are low-resolution approximations of the solution fields acquired by sampling the constitutive field, $\boldsymbol{m}(x)$, at a resolution $M^3$ (with $M<<N$) and performing low-cost, low-resolution DNS simulations.
$D(\cdot)$ is a downsampling function mapping from $N^3$ to $M^3$; in this paper, a windowed average.
$L(\cdot)$ is a selected loss function, see the \methodsref{sec:methods} for more details.
}

{
A purely data-driven learning problem is expected to produce a poorly trained neural surrogate because the low-resolution supervision is missing important information at higher resolution.
Specifically, it is unable to inform any high-resolution frequency content present in high-resolution predictions: for instance when the resolution factor $M/N=2$ in 3D, the low-resolution supervision contains just one eighth of the high-resolution frequency content.
As demonstrated later in this section, for micromechanics problems, this information is absolutely necessary for high-fidelity predictions.
}

\begin{figure*}[h!]
  \centering
  \includegraphics[width=1.0\textwidth]{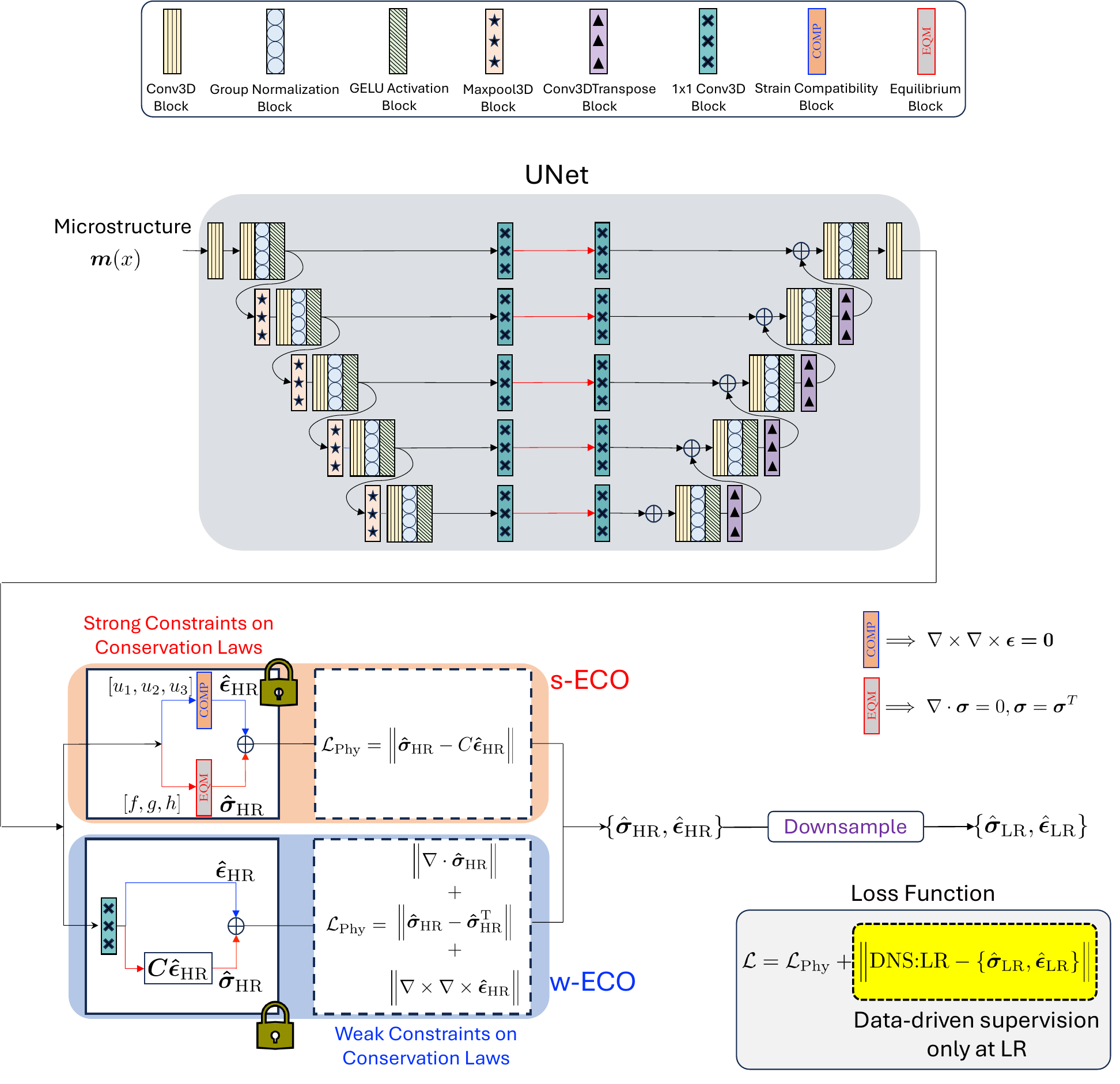}
  \caption{\textbf{Architecture.}
  Physics-based neural super-resolution is performed with a UNet backbone implemented with the proposed ECO formulation. 
  The UNet takes the microstructure $\boldsymbol{m}(x)$ as its input.
  s-ECO strongly enforces the conservation laws through the strain compatibility and equilibrium blocks while weakly imposing the constitutive law.
  w-ECO, on the other hand, strongly enforces the constitutive law and weakly imposes the conservation law via its loss function.
  Data-driven supervision is performed only at low resolution (LR), aided by coarser and cheaper simulations (DNS:LR).
  }
  \label{fig:architecture}
\end{figure*}

{
We propose a physics-based neural super-resolution framework, see \autoref{fig:framework_overview}, that trains high-resolution neural surrogates by supplementing the low-resolution training data using the targeted system physics (e.g., the governing PDEs and constitutive equations) to account for the missing information.
We propose and utilize the Equilibrium Conserving Operator (ECO) formulation to achieve this by strongly incorporating the system physics into the neural surrogate's architecture, \autoref{fig:architecture}.
We direct the interested reader to the \methodsref{sec:methods} for a complete derivation of the produced super-resolution framework. We formulate this expanded ECO-based learning problem for small strain elasticity under periodic boundary conditions, but we apply the same framework to finite strain, non-periodic problems.
In the two coming experiments, we considered two implementations using different types of Equilibrium Conserving Operator (ECO) neural architecture: a strong formulation (s-ECO) in which we implement conservation laws through network architectural blocks that strictly enforce stress equilibrium and deformation compatibility, and a weak formulation (w-ECO) in which the conservation laws are accounted for in the loss function using standard physics-informed neural network (PINN) methods.
This choice critically impacts implementation flexibility and physical plausibility.
The integration of constitutive laws offers greater flexibility in capturing complex, nonlinear material behaviors but may risk producing physically implausible results without proper constraints.
Conversely, encoding conservation laws prevents unphysical solutions, but increases architectural complexity.
}

\subsection*{Embedded Pores}
\begin{figure*}[h!]
  \centering
  \includegraphics[width=1.0\textwidth]{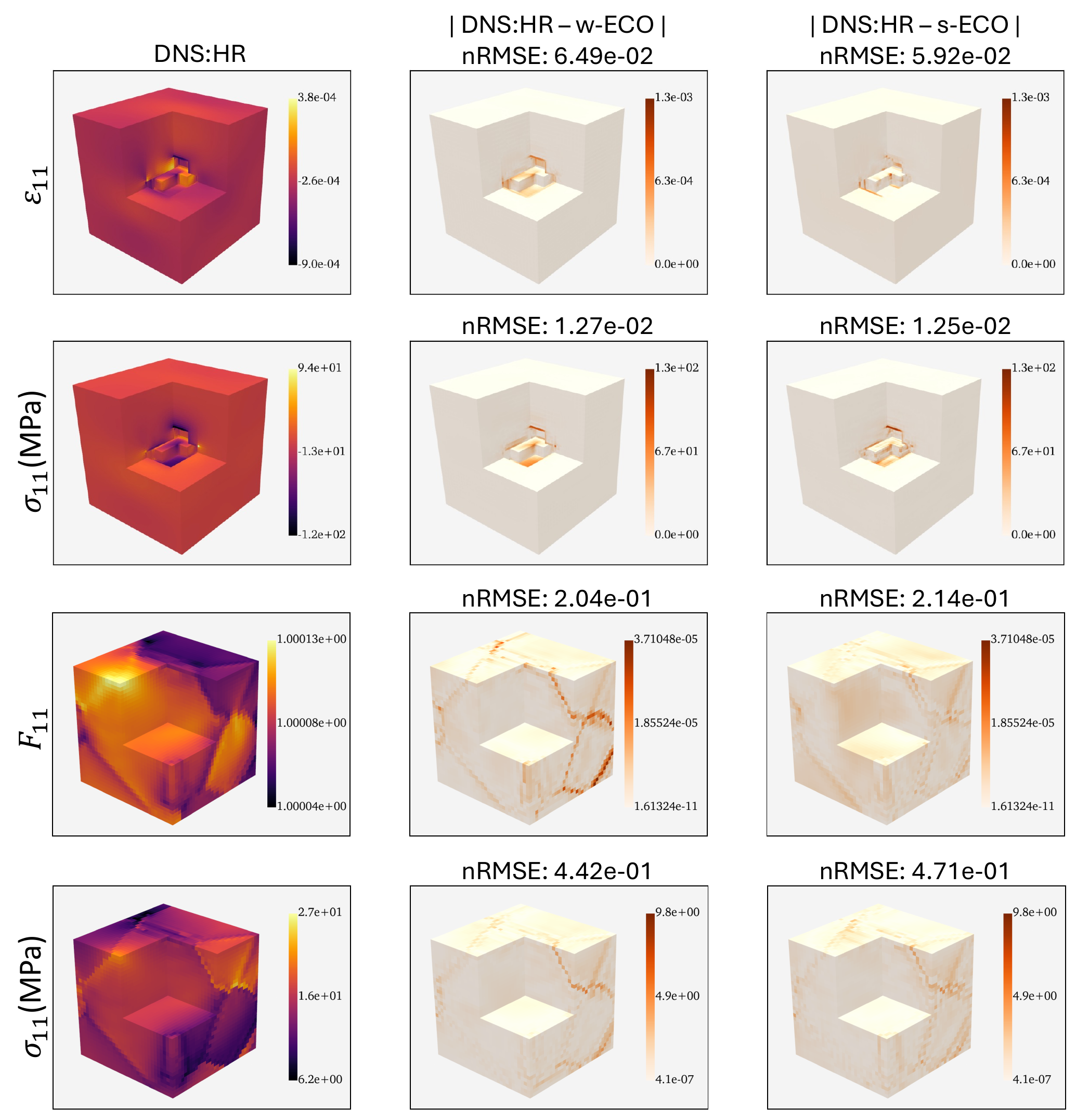}
  \caption{\textbf{Strains and stresses predictions.}
  We compare the material response predicted by the physics-based neural super-resolution implementations -- w-ECO (center column) \& s-ECO (right column) -- with the DNS (left column) at twice the resolution of the training data. 
  Top two rows compares the stresses and elastic strains in the embedded pores case study. 
  Bottom two rows compares the stresses and elastic deformation gradients in the polycrystalline microstructures case study.}
  \label{fig:combined_stress_strain}
\end{figure*}

{
The top two rows of \autoref{fig:combined_stress_strain} show the high-resolution ($256^3$) DNS numerical solutions for the embedded pores case study (left column) contrasted with the absolute error in the predictions made by the two implementations (w-ECO: center column; s-ECO: right column) of our physics-based neural super-resolution framework.
For both implementations, qualitatively, the fields show good agreement with the DNS results.
The physics-based supervision effectively incorporates high-resolution details and eliminates any multi-resolution artifacts that could occur with purely data-driven supervision.
Notably, this agreement is achieved \textit{without} using any high-resolution simulations during training.
Furthermore, the predictions exhibit acceptably low mean errors, as tabulated in \autoref{tab:errors}, except near the phase boundaries, which are challenging to model due to large gradients.
In terms of numerical error, measured using the normalized root mean squared error (nRMSE) against the high-resolution numerical simulation, both implementations perform similarly, with s-ECO slightly outperforming the w-ECO.
}

\begin{table*}[h!]
    \centering
    \caption{nRMSE for w-ECO and s-ECO across the test dataset for the embedded pore case study (column 2 and 3) and the test dataset for the  polycrystalline microstructures case study (column 5 and 6).}
    \label{tab:errors}

    \begin{tabular}{|ccc|ccc|}
        \toprule
        & \multicolumn{2}{c|}{\bf Embedded Pores} & & \multicolumn{2}{c|}{\bf Polycrystalline Microstructures}\\
        \midrule
         & {\bf w-ECO} & {\bf s-ECO} & & {\bf w-ECO} & {\bf s-ECO}\\
        {\bf Strains} &  & & {\bf Strains} &  &\\
        \midrule
        $\epsilon_{11}$ & 6.43e-02 $\pm$ 5.23e-04 & 5.92e-02 $\pm$ 3.14e-04 & $F^e_{11}$ & 2.04e-1 $\pm$ 1.52e-2 & 1.82e-1 $\pm$ 3.57e-2 \\
        $\epsilon_{12}$ & 1.67e-02 $\pm$ 3.21e-03 & 1.62e-02 $\pm$ 2.79e-03 & $F^e_{12}$ & 2.66e-1 $\pm$ 4.74e-2 & 2.44e-1 $\pm$ 2.87e-2 \\
        $\epsilon_{13}$ & 8.45e-03 $\pm$ 1.51e-03 & 9.19e-03 $\pm$ 1.80e-03 & $F^e_{13}$ & 2.67e-1 $\pm$ 4.14e-2 & 3.06e-1 $\pm$ 2.27e-2 \\
        $\epsilon_{22}$ & 6.50e-02 $\pm$ 6.06e-04 & 5.98e-02 $\pm$ 3.21e-04 & $F^e_{22}$ & 1.93e-1 $\pm$ 2.16e-2 & 1.96e-1 $\pm$ 1.90e-2 \\
        $\epsilon_{23}$ & 9.74e-03 $\pm$ 1.83e-03 & 1.04e-02 $\pm$ 1.90e-03 & $F^e_{23}$ & 3.95e-1 $\pm$ 7.83e-2 & 3.77e-1 $\pm$ 1.03e-1 \\
        $\epsilon_{33}$ & 7.77e-02 $\pm$ 4.43e-04 & 7.87e-02 $\pm$ 1.52e-04 & $F^e_{33}$ & 1.99e-1 $\pm$ 2.71e-2 & 1.97e-1 $\pm$ 2.17e-2 \\
        -- & -- & -- & $F^e_{21}$ & 3.26e-1 $\pm$ 3.72e-2 & 3.00e-1 $\pm$ 3.41e-2 \\
        -- & -- & -- & $F^e_{31}$ & 3.44e-1 $\pm$ 3.12e-2 & 3.38e-1 $\pm$ 6.67e-2 \\
        -- & -- & -- & $F^e_{32}$ & 3.93e-1 $\pm$ 8.15e-2 & 2.96e-1 $\pm$ 6.88e-2 \\
        \midrule
        {\bf Stresses} & &  &  {\bf Stresses} &  &\\
        \midrule
        $\sigma_{11}$ & 9.39e-03 $\pm$ 2.43e-03 & 1.08e-02 $\pm$ 1.76e-03 & $\sigma_{11}$ & 4.42e-1 $\pm$ 2.46e-2 & 4.94e-1 $\pm$ 4.15e-2 \\
        $\sigma_{12}$ & 1.67e-02 $\pm$ 3.21e-03 & 1.64e-02 $\pm$ 3.28e-03 & $\sigma_{12}$ & 2.58e-1 $\pm$ 2.94e-2 & 2.86e-1 $\pm$ 1.11e-2 \\
        $\sigma_{13}$ & 8.45e-03 $\pm$ 1.51e-03 & 8.10e-03 $\pm$ 1.64e-03 & $\sigma_{13}$ & 2.77e-1 $\pm$ 4.93e-2 & 3.05e-1 $\pm$ 2.27e-2 \\
        $\sigma_{22}$ & 8.98e-03 $\pm$ 2.33e-03 & 1.02e-02 $\pm$ 1.56e-03 & $\sigma_{22}$ & 3.97e-1 $\pm$ 3.31e-2 & 4.22e-1 $\pm$ 1.75e-2 \\
        $\sigma_{23}$ & 9.73e-03 $\pm$ 1.83e-03 & 9.21e-03 $\pm$ 1.80e-03 & $\sigma_{23}$ & 2.84e-1 $\pm$ 2.78e-2 & 2.96e-1 $\pm$ 1.94e-2 \\
        $\sigma_{33}$ & 7.64e-02 $\pm$ 4.46e-04 & 7.45e-02 $\pm$ 4.71e-04 & $\sigma_{33}$ & 4.03e-1 $\pm$ 3.25e-2 & 4.31e-1 $\pm$ 2.31e-2 \\
        \bottomrule
    \end{tabular}
\end{table*}

\begin{figure*}[h!]
  \centering
  \includegraphics[width=1.0\linewidth]{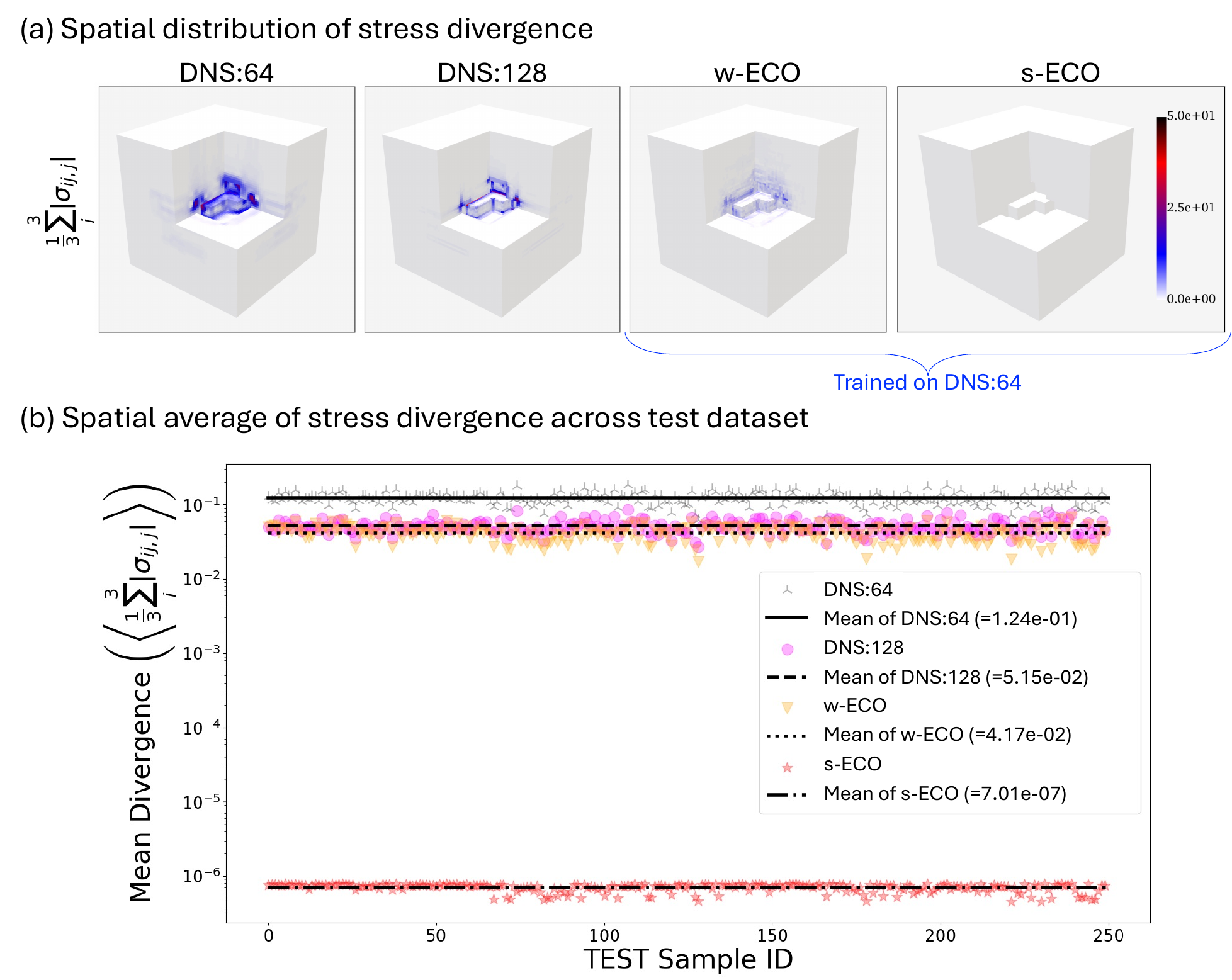}
  \caption{\textbf{Stress divergence.}
  (a) Spatial distribution of stress divergence for a sample randomly selected from the test dataset for low- (DNS:64) and high-resolution (DNS:128) direct numerical simulation (DNS) results and compared to predictions from the (weakly enforced) physics-informed equilibrium conserving operator (w-ECO) and the strong ECO (s-ECO).
  (b) Spatial average of stress divergence across all 250 samples in the test dataset.}
  \label{fig:pore_div}
\end{figure*}

{
The first significant difference between the two implementations is evident when examining how well the predictions align with the actual distribution of mechanical fields.
\autoref{fig:pore_div} compares the divergence of the predicted stress fields, $\nabla \cdot \boldsymbol{\sigma}$, from w-ECO and s-ECO with various DNS resolutions.
The importance of the s-ECO's equilibrium block is clear; it achieves near machine precision divergence\footnote{As is standard in machine-learning applications, we used $\mathrm{fp}32$. The increase from $1\times 10^{-8}$ arises during rescaling.}.
In contrast, while w-ECO reduces divergence compared to the training data and provides necessary physics supervision for super-resolution, it does not match the s-ECO's adherence to the conservation laws.
Reducing divergence-free error further with only a softly constrained PDE loss is challenging due to the slow training dynamics of such scientific machine-learning algorithms \cite{wang2022and}.
Notably, neither of the DNS solutions (i.e., the DNS:LR used for training or the DNS:HR used for testing) achieves the same level of compliance with mechanical equilibrium as s-ECO\footnote{This arises because the FTT solver implementation is strain mediated.
Solution iterations are stopped when the relative magnitude of updates in the predicted strain fields goes below a set threshold.}.
On one hand, this difference is important when interpreting the previously discussed loss measures.
s-ECO predicts fields which, by design, lie within the strict solution set of the system's PDEs, representing a different, albeit nearby, function space than the training data.
Therefore, measuring success solely by loss minimization is insufficient.
On the other hand, w-ECO strongly constrains the surrogate network to satisfy the constitutive law, which is expected to outperform s-ECO in that regard.
We defer the development of an architecture that strongly satisfies both the PDEs and the constitutive relationships to future work.
}

\begin{figure*}[h!]
  \centering
  \includegraphics[width=0.98\linewidth]{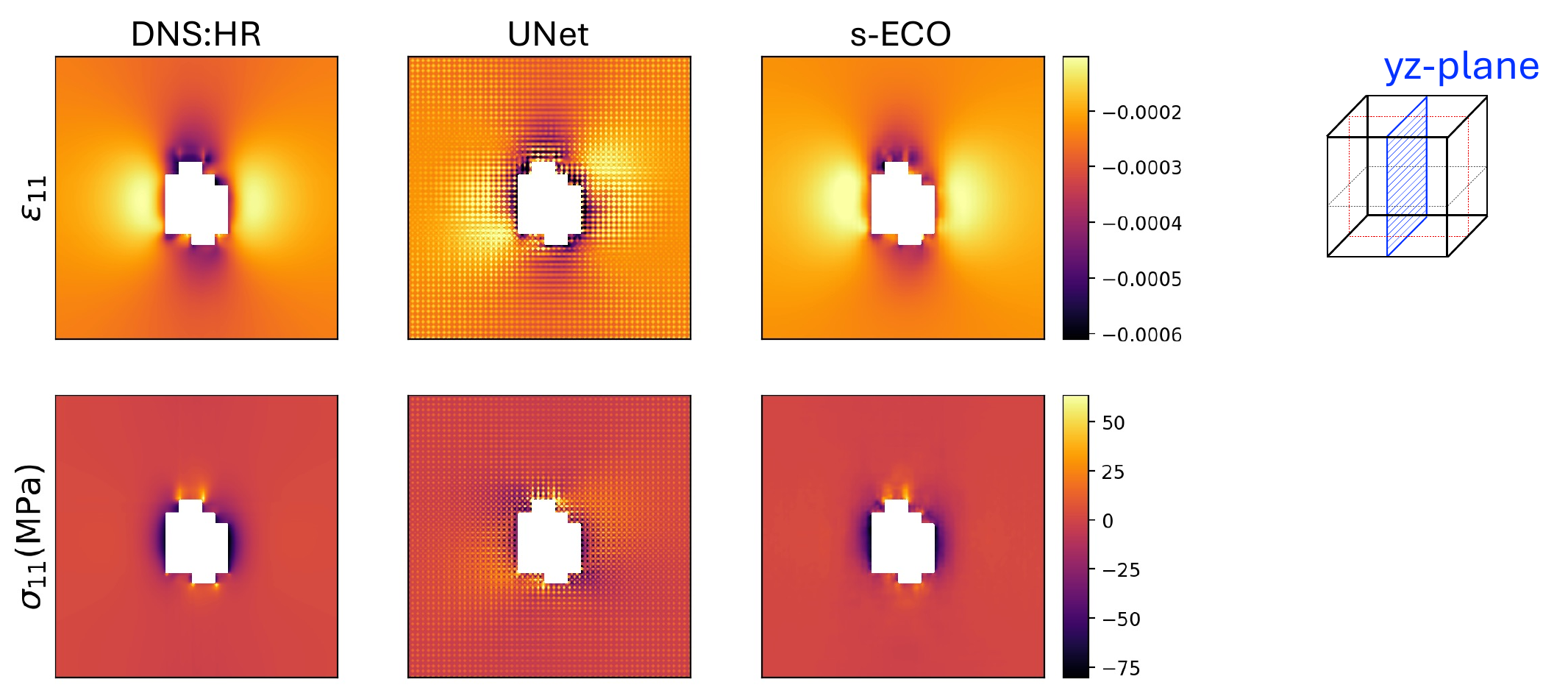}
  \caption{\textbf{Importance of incorporating physics in physics-based neural super-resolution.}
  We compare the high-resolution (HR) material response, for the embedded pore case study, predicted by data-driven UNet (middle column) and the present s-ECO (right column) with the HR reference simulated using FFT solver (left column, DNS:HR). 
  Note that the neural surrogates were trained on cheaper DNS low resolution whose resolution was half of the DNS:HR test dataset.
  Embedding the physics enables s-ECO to avoid the non-physical artifacts present in the standard UNet prediction.}
  \label{fig:nn_vs_spnn}
\end{figure*}

{
We also examined the significance of incorporating physics into the physics-based super-resolution framework by comparing a standard UNet with s-ECO.
The standard UNet is trained in the same multi-resolution setting using only data-driven supervision at the low resolution and without any physics-based supervision.
In \autoref{fig:nn_vs_spnn}, we compare the high-resolution elastic strains and stresses estimated by both the UNet and s-ECO against the the DNS FFT solver (DNS:HR).
While the UNet achieves low training and validation errors at the low resolution, it produces non-physical artifacts in the high-resolution predictions.
In contrast, s-ECO does not exhibit such artifacts.
This dichotomy is consistent across all components; see extended analysis in the \autoref{app:extendedanalysis} for a comparison of the remaining tensor components.
By strongly enforcing the PDEs in its architecture and softly constraining the constitutive laws in the training loss, s-ECO eliminates the artifacts predicted by the UNet.
We also attempted mechanical super-resolution using a standard deep operator network (DeepONet) \cite{lu2021learning} and Tensorized Fourier Neural Operator (TFNO) \cite{kossaifi2023multi} architectures instead of the UNet.
However, despite extensive hyperparameter optimization, neither DeepONet nor TFNO training losses converged.
We believe such a behavior is a result of very high gradients in the elastic strains and stresses near the embedded pores.
This observation aligns with other studies \cite{gupta2022towards, ovadia2023ditto} that report greater robustness of UNet-based architectures to high spatial gradients.
In the next case study, we explore additional perspectives to understand the effectiveness of the proposed super-resolution framework.
}

\subsection*{Polycrystalline Microstructures}
\label{sec:results_polycrystal}
{
The second case study demonstrates the applicability of our physics-based neural super-resolution framework to polycrystalline systems and contrasts it with an alternative existing solution: operator learning methods.
Additionally, it explores the implications of the differences between selected implementations in greater detail.
For polycrystalline systems, we found model performance to be extremely sensitive to microstructure representation (see section on representation in the \autoref{app:polycrystalline_inputs}).
}

{
Starting with the low-resolution to high-resolution super-resolution problem, the results align with those from the embedded pore case study.
The bottom two rows of \autoref{fig:combined_stress_strain} show example predictions of the $\sigma_{11}$ and $F_{11}^e$ fields, which are the most dominant fields due to the tensile $x$-direction loading condition.
As in the previous section, the s-ECO's high-resolution predictions effectively resolve the key features in the stress and deformation gradient fields.
Again, the losses are similar; this time w-ECO slightly outperforms s-ECO, as shown in \autoref{tab:errors}.
Both models exhibit slight relative increases in the off-diagonal, non-hydrostatic terms.
However, despite these similarities, the spatial distributions of the losses differ: the w-ECO's loss is primarily concentrated at the grain boundaries, whereas s-ECO distributes the loss more evenly across the domain.
This consistent difference is likely related to the architecture design discussed earlier.
Like in the embedded pore case study, s-ECO achieves near machine precision adherence to the governing partial differential equations.
}

{
Alternative strategies for super-resolution exist in the scientific machine-learning literature.
As an external benchmark, we compared our physics-based super-resolution framework against operator learning methods \cite{li2020fourier, kossaifi2023multi}.
These methods claim to have solved the problem by learning mesh-independent function-to-function mappings.
However, we argue that mesh independence does not necessarily imply super-resolution for micromechanical problems\footnote{This underappreciated fact has been abstractly argued previously. In a more general context, Raonic \textit{et al.} argue that operator methods are limited to mesh-independence on band-limited functions \cite{raonic2023convolutionalneuraloperator}. Micromechanical problems are inherently not band-limited due to the sharp grain and phase boundaries in materials, leading to oversmoothing and frequency artifacts.}.
}

{
As such, for comparison, we trained a TFNO network \cite{kossaifi2023multi} in a traditional data-driven setting \cite{li2020fourier}.
This network was trained to predict low-resolution stress and deformation gradient fields from low-resolution microstructures as input.
Subsequently, it was applied to high-resolution microstructures during inference by leveraging its mesh independence.
We used the same training hyperparameters identified for s-ECO (see Method)\footnote{The maximum learning rate was independently adjusted, but we found the same optimum as Optuna.}.
The network architecture hyperparameters were optimized through a simple grid search.
%
}

\begin{figure*}[h!]
  \centering
  \includegraphics[width=1.0\textwidth]{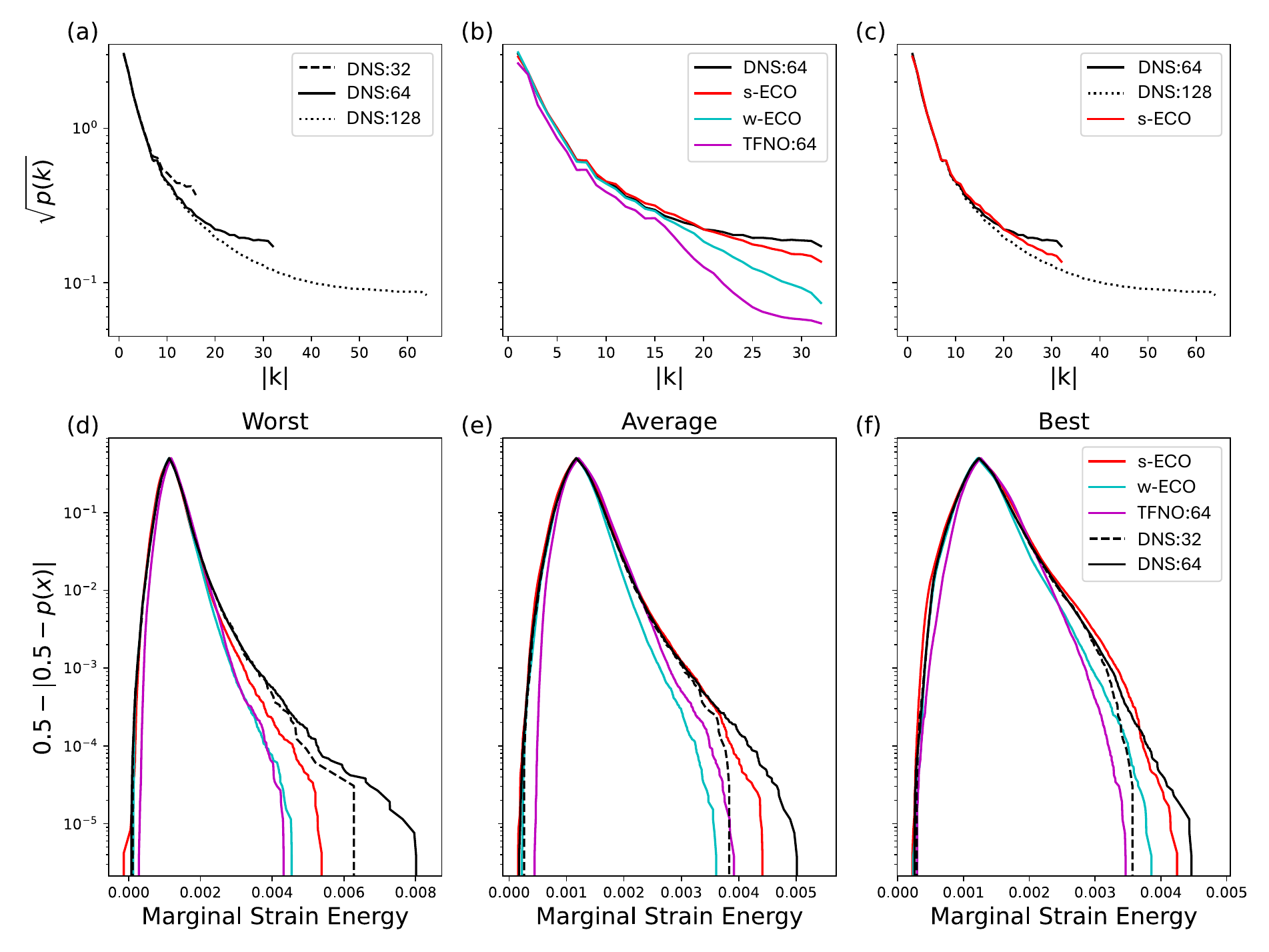}
  \caption{\textbf{Quantitative comparison of the super-resolution performance between the s-ECO implementation and an operator learning method (TFNO).}
  (a)-(c) Radially averaged power spectra of the $\sigma_{11}$ stress component from a representative test simulation.
  (d)-(f) Marginal distribution of the elastic strain energy density for three representative test simulations. The marginal distribution is visualized to highlight differences in the tails (i.e. the extreme values).}
\label{fig:discussion_combined_fouriermarginal}
\end{figure*}

{
The missing information in the low-frequency data is organized compactly in the frequency domain.
\autoref{fig:discussion_combined_fouriermarginal}(a)-(c) compares the frequency content between the DNS simulation at multiple resolutions, the TFNO, and the two physics-based implementations.
This comparison is visualized in 1D by spherically averaging the frequencies.
In 3D, the low-resolution simulations are missing $7/8$ of the total frequency information, estimating only the frequencies below half of the maximum high-resolution frequency ($|k|<16$), see \autoref{fig:discussion_combined_fouriermarginal}(a).
This analysis highlights the major strength of the s-ECO implementation: its ability to effectively recover missing high-frequency information in physics-based neural super-resolution, as shown in \autoref{fig:discussion_combined_fouriermarginal}(b).
The s-ECO's $64$-resolution predictions remain in good accordance with the DNS:$64$ results, even beyond the frequencies present in the low-resolution training data.
Although a slight divergence occurs at high resolutions, it corresponds to the frequency information in the higher-resolution ground truth (DNS:$128$), as illustrated in \autoref{fig:discussion_combined_fouriermarginal}(c).
This trend is consistent across all stress components.
}

{
In contrast, the w-ECO implementation underestimates high frequencies but still outperforms the TFNO, demonstrating the value of the physics-based super-resolution framework, even with suboptimal implementation.
The TFNO drastically underestimates the high frequencies, showing good agreement until $|k|=15$ (the maximum frequency available from the low-resolution simulations), after which it sharply deviates from expectation.
This behavior is not surprising, as band-limited predictions \cite{raonic2023convolutionalneuraloperator} and oversmoothing \cite{oommen2024integrating} are well documented behaviors of operator methods.
Clearly, the TFNO would benefit from training using the proposed framework rather than using standard operator methods.
}

{
During hyperparameter optimization for the TFNO, we found that improved performance on the training problem (mapping low-resolution microstructures to stress and strain) did not correlate with better super-resolution predictions.
\autoref{fig:discussion_combined_fouriermarginal}(b) shows the TFNO trained with optimal hyperparameters for the low-resolution problem, consistent with previous recommendations for operator learning implementations \cite{kossaifi2023multi}.
Despite this, none of the hyperparameters tested successfully matched the high-frequency content of the high-resolution simulations, and the reported performance trends were consistent across the entire dataset.
}

\begin{figure*}[h!]
  \centering
  \includegraphics[width=1.0\textwidth]{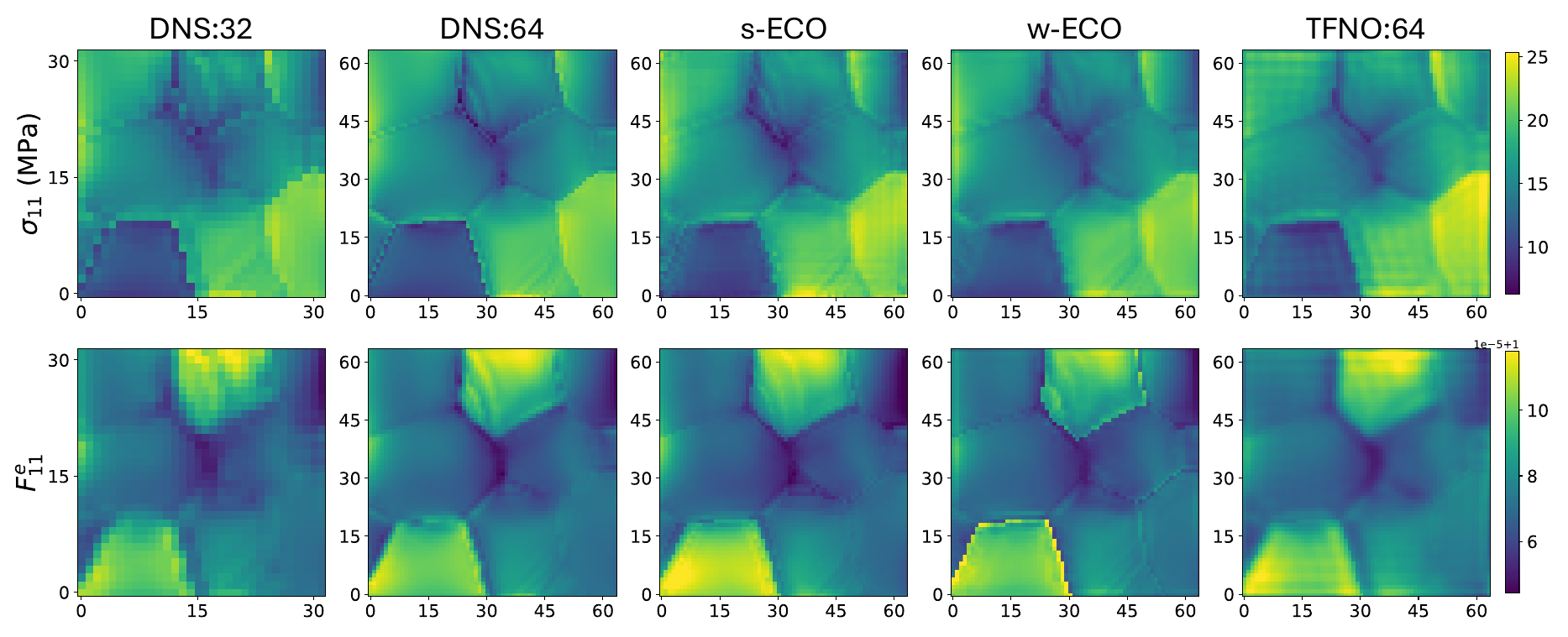}
  \caption{\textbf{Distribution of local fields: a comparison between s-ECO and TFNO.}
  Comparison of single component, 2D slices (YZ-slice along $x=32$) of stress ($\sigma_{11}$, top row) and elastic deformation gradient ($F^e_{11}$, bottom row) fields between the direct numerical simulation ($\mathrm{DNS}:32$ -- training resolution -- and $\mathrm{DNS}:64$, first two columns) and machine-learned predictions (s-ECO (third column), w-ECO (fourth column), and TFNO (last column)).}
  \label{fig:discussion_grainboundary}
\end{figure*}

{
The observed mathematical oversmoothing in w-ECO and TFNO approaches manifests as extreme value errors and grain-boundary errors, respectively.
\autoref{fig:discussion_grainboundary} compares a 2D slice of the predicted stress and elastic deformation gradient fields from the three approaches against the DNS (both at low resolution (DNS:32) for training and high resolution (DNS:64) for testing).
Both physics-based super-resolution implementations extend beyond the information available in the low-resolution training data (DNS:32).
In particular, s-ECO effectively super-resolves grain boundaries and captures extreme-value stresses, successfully resolving for instance stresses in the central grain boundary (bottom axis = 35, left axis = 35) and the large deformation in the top center grain (bottom axis = 35, left axis = 60).
In contrast, w-ECO predictions exhibit undersmoothing within grains, while grain boundaries are oversharpened.
The TFNO shows various errors, including Gibbs-like oscillations and significant grain-boundary smoothing, with inconsistent error modalities -- displaying both oversharpened and undersharpened grain boundaries compared to the ground truth.
This inconsistency suggests instability due to a lack of high-frequency information during training.
Additionally, we note that s-ECO performs better in predicting deformation gradients than stress fields.
We hypothesize that the deformation gradients are easier to learn because the compatibility block contains only one finite difference derivative.
Exploring alternative implementations of the s-ECO's stress equilibrium block presents a potential area for future improvement.
}

{
In addition to matching frequency content, another practically important measure of super-resolution performance is the prediction of field concentrations, particularly extreme elastic strain energies, which drive many nonlinear processes such as dislocation motion or fatigue \cite{mcdowell_strainenergy, przybyla_strainenergy}.
Specifically, \autoref{fig:discussion_combined_fouriermarginal}(d)-(f) compares the marginal distribution of elastic strain energy for three high-resolution test cases:
a poor example -- panel (d),
an average example -- panel (e), and
an ideal example -- panel (f).
We plot $0.5 - |0.5 - p(x)|$ to emphasize the quality of predictions at extreme values (i.e., the low probability tails).
A performance hierarchy emerges consistent with our previous discussions.
On average, s-ECO successfully extrapolates beyond the low-resolution limit, leveraging the physics of the system to fill in high-resolution information.
In contrast, both the w-ECO and TFNO systematically underpredict these critical tail values across all examples in the testing dataset.
In the ideal example in panel (f), s-ECO successfully recreates the high-resolution prediction.
Comparing the skew of the marginal distribution across the three examples reveals a persistent indicator of poor performance: all models struggle with high peak elastic strain energy, likely due to oversmoothing in the selected neural network architecture.
Previous work by Kelly and Kalidindi has shown that iterative architectures outperform traditional one-shot architectures in handling extreme values in single-resolution applications \cite{kelly2024therino, kelly2021recurrent}.
We plan to explore this type of architecture in future work.
Overall, the present physics-based super-resolution framework exceeds the information content of the low-resolution simulations it was trained on, effectively extrapolating remaining frequency content using system physics.
}

\begin{figure*}[h!]
  \centering
  \includegraphics[width=0.7\linewidth]{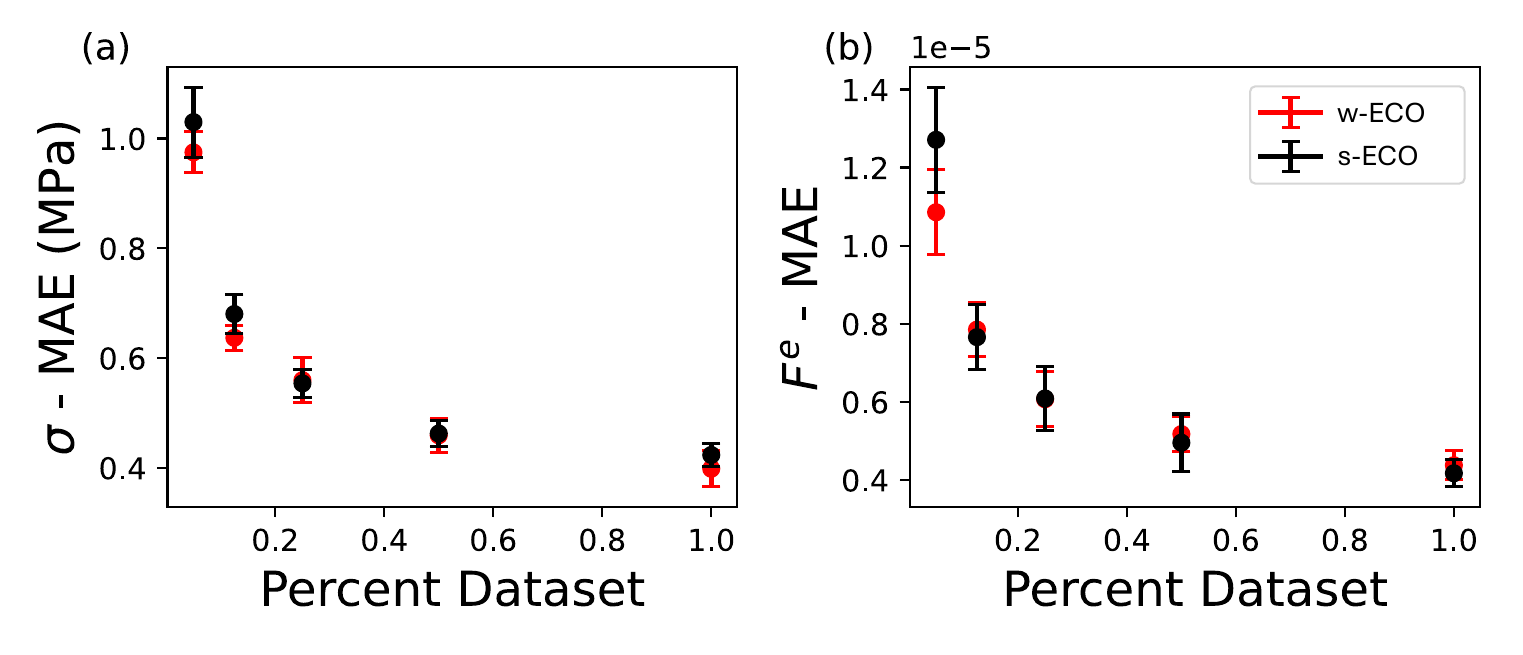}
  \caption{\textbf{Comparison of the s-ECO and w-ECO model performance as the quantity of training data is reduced.} The complete dataset includes $784$ training tuples.}
  \label{fig:pc_datascaling}
\end{figure*}

{
To conclude this section, we examine the training stability of the proposed framework with decreasing training data.
\autoref{fig:pc_datascaling} shows the framework's performance scaling with the amount of collected data\footnote{In this comparison, the number of steps (i.e. gradient updates) is kept constant as the amount of data is reduced. To achieve this, the number of epochs is increased. This leads to each data point being observed an increased number of times to make up for the decreased amount of data. We used the established hyperparameters for all experiments. We do not use early stopping.}.
The full dataset consists of $783$ low-resolution simulations.
Both implementations demonstrate excellent stability, with performance remaining largely unchanged despite significant reductions in training data.
We hypothesize that this stability arises from the physics-based supervision integrated into the framework \cite{wang2021learning, karniadakis2021physics}.
}

\subsection*{Ultra-High Resolution Evaluation}
\begin{figure*}[h!]
  \centering
  \includegraphics[width=1.0\textwidth]{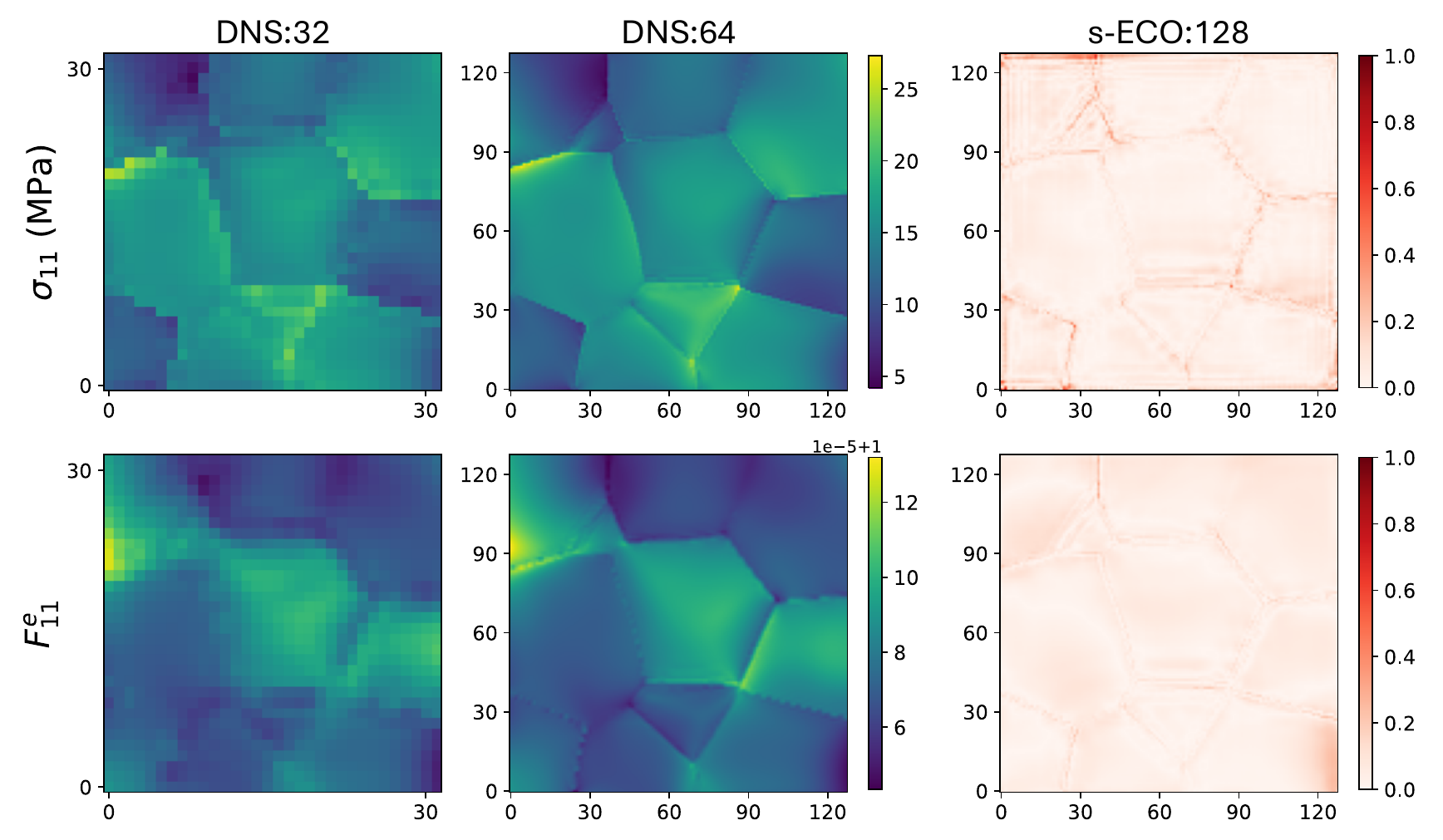}
  \caption{\textbf{Performance remains stable even with greater resolution gaps}. Comparison of an example $\sigma_{11}$ slice from the low resolution DNS (similar to the data used for training), a ultra-high resolution DNS, and predictions from a ultra-high resolution s-ECO surrogate. s-ECO predictions are displayed in relative absolute error.}
  \label{fig:discussion_ultrarez}
\end{figure*}

{
In the previous case studies, we demonstrated the model's performance on DNS at double the training resolution.
To further test the framework's capabilities, we trained an ultra-high resolution surrogate at $128^3$ using low-resolution data ($32^3$).
We maintained the same training and architecture hyperparameters as before.
\autoref{fig:discussion_ultrarez} compares slices from the low-resolution testing (DNS:32), ultra-high resolution testing (DNS:128), and the s-ECO's predictions (s-ECO:128).
s-ECO effectively sharpens grain boundaries and reveals distinct behaviors in previously obscured grains (e.g., the grain around $35$ on the horizontal axis and $100$ on the vertical).
The surrogate model predicts the high-resolution field with good fidelity, with relative absolute errors generally below $20\%$ (measured in relative absolute error).
High errors to this level are primarily restricted to the grain and domain boundaries.
These results indicate that the framework remains stable even with larger resolution jumps.
}

\section*{Discussion}
\label{sec:discussion}

{
Our results demonstrate the ability of the ECO-based super-resolution framework to train high-resolution neural surrogates stably, even with significant differences between training and inference resolutions.
 This, combined with its stability when reducing training dataset size significantly lowers the upfront training costs.
 In the polycrystalline microstructures case study the DNS simulation cost increased by $330\%$ from $32^3$ to $64^3$ and by an additional $2\,500\%$ from $64^3$ to $128^3$, resulting in an overall increase of $8\,330\%$ between $32^3$ and $128^3$.
 Consequently, the surrogate is nearly two orders of magnitude cheaper to train than a direct implementation requiring high-resolution training data.
 To put this into perspective, the $1\,000$ simulations used for training and validating the model took comparable simulation time to the $5$ ultra-high resolution simulations performed for testing.
 This $83.3$ times reduction in cost is likely conservative; it assumes that a surrogate trained on high resolution data would display similar data scaling, see \autoref{fig:pc_datascaling}. Even greater reduction is achievable if this is not true. 
Briefly, a second significant source of computational cost for machine-learning surrogates is training time.
The analysis in mixed-precision acceleration in the \autoref{app:mixed_precision} demonstrates that ECO and the proposed physics-based neural super-resolution can be stably combined with standard acceleration methods like mixed precision training to combat this cost as well.
 Altogether, this significant reduction in the upfront, one time data collection cost eliminates one significant source of the computational divide between numerical simulations and machine learning.
}

{
The benefits of this framework are expected to compound when applied to nonlinear constitutive laws and dynamic problems, given the substantial increase in computational costs associated with DNS simulations in these contexts.
Integrating nonlinear constitutive laws into the s-ECO's loss function presents challenges that warrant further investigation in future work.
}

{
In summary, in this paper, we introduce an Equilibirium Conserving Operator (ECO) formulation for strongly incorporating the governing physics into neural networks for micromechanics problems. 
We utilize ECO to build a physics-based neural super-resolution framework for solving solid mechanics problems.
This framework facilitates the training of high-resolution neural network models using only computationally inexpensive, low-resolution data.
It relies on ECO to supplement missing high-resolution information in the low-resolution data using governing physics.
The ECO architecture consists of a strongly physics-informed U-shaped neural network (s-ECO) which incorporates conservation laws directly.
Such architecture ensures mechanical equilibrium and deformation compatibility via an `equilibrium' block, which strongly enforces the conservation of linear (i.e. $\nabla \cdot \sigma=0$) and angular (i.e. $\sigma = \sigma^T$) momentum and a `compatibility' block, which enforces compatibility of the deformation field.
The framework trains a high-resolution s-ECO with supervision from the low-resolution data and a soft penalty on the constitutive law.
Results from case studies on microstructures with embedded pores and polycrystalline materials demonstrate that s-ECO outperforms traditional weakly enforced implementations and mesh-free operator methods in predicting high-resolution mechanical fields and recovers missing high-frequency content.
The ability to train on low-resolution data makes such neural surrogate nearly two orders of magnitude cheaper to train than direct implementations requiring high-resolution training data.
This approach effectively eliminates the substantial upfront data collection cost traditionally associated with neural network methods in this domain and is readily generalizable across various applications.
}

\section*{Methods}
\label{sec:methods}


\subsection*{Problem setup}
{
Mathematically, for an elastic solid, the specific set of parametric PDEs considered are:
}
\begin{align}
    f: \boldsymbol{C}(\boldsymbol{m}(\boldsymbol{x})), \bar{\boldsymbol{\epsilon}} \mapsto \boldsymbol{\sigma}&(\boldsymbol{x}), \boldsymbol{\epsilon}(\boldsymbol{x}) \\
    \textrm{s.t.} \ \ \ \ \ \ \ \ \ \nabla \cdot \boldsymbol{\sigma}(\boldsymbol{x}) &= \boldsymbol{0}\label{eq:cons1} \\
    \nabla \times \nabla \times \boldsymbol{\epsilon}(\boldsymbol{x} ) &= \boldsymbol{0} \label{eq:kinematic_compatibility}\\
    \boldsymbol{\sigma}(\boldsymbol{x})^T &= \boldsymbol{\sigma}(\boldsymbol{x})\label{eq:stress_sym} \\
    \left \langle \boldsymbol{\epsilon}(\boldsymbol{x}) \right \rangle &= \bar{\boldsymbol{\epsilon}} \\
    \boldsymbol{C}(\boldsymbol{m}(\boldsymbol{x})) \boldsymbol{\epsilon}(\boldsymbol{x}) &= \boldsymbol{\sigma}(\boldsymbol{x})\label{eq:const} \\
    \boldsymbol{\sigma}(\boldsymbol{x}), \boldsymbol{\epsilon}(\boldsymbol{x}) \mathrm{\ ar}&\mathrm{e \ periodic.}
\end{align}
{
\noindent Here, $\boldsymbol{x}\in\Omega\subset \mathbb{R}^3$, 
$\boldsymbol{m}(\boldsymbol{x})$ is the material microstructure function, 
$\boldsymbol{C}(\boldsymbol{m}(\boldsymbol{x}))$ is the fourth-rank elastic stiffness tensor defined by the material and its microstructure, and 
$\langle \cdot \rangle$ defines a spatial average.
For our purpose, the uniaxial deformation boundary condition $\bar{\boldsymbol{\epsilon}}$ is kept constant and is included only implicitly.
In practice, the spatial domain and the corresponding mechanical fields are discretely sampled to facilitate solving via a numerical scheme.
In the above, \autoref{eq:cons1} and \autoref{eq:kinematic_compatibility} are the conservation laws imposed over the entire domain, while \autoref{eq:const} is the constitutive law defining the local material behavior (linear in this case).
}

\subsection*{Implementation: strong Equilibrium Conserving Operator}
{
We introduce a strong physics-informed equilibrium
conserving operator (denoted as s-ECO) for solid mechanics, designed as a practical tool for implementing the governing equations listed in \autoref{eq:cons1} -- \autoref{eq:const} and in \autoref{eq:mathform}.
The proposed s-ECO architecture embeds the governing partial differential equations directly into the network architecture, ensuring that any fields predicted by the network are strongly conformant with the governing conservation laws.
This architecture is inspired by structure-preserving designs in dynamics \cite{zhang2022gfinns} and in fluid dynamic PDEs \cite{mohan2023embedding}.
As illustrated in \autoref{fig:architecture}, to achieve built-in conservation laws, the network architecture includes two key architectural blocks:
(1) the `compatibility' block and
(2) the `equilibrium' block.
}

{
The compatibility block enforces the kinematic compatibility criterion (i.e., \autoref{eq:kinematic_compatibility}), ensuring that the estimated kinematic strain tensors are derived from valid displacement vectors.
For the infinitesimal strain problem, we utilize \autoref{eq:strain_compatibility} to enforce compatibility.
This block takes a vector field, $\boldsymbol{u}(\boldsymbol{x}) \in \mathbb{R}^3$ for $\boldsymbol{x} \in \Omega \subset \mathbb{R}^3$, and outputs a compatible kinematic strain tensor field.
A similar expression is used for elastic deformation gradients.
}

\begin{equation}
    \epsilon_{ij} = \frac{1}{2}(u_{i,j} + u_{j,i}).
    \label{eq:strain_compatibility}
\end{equation}

{
The equilibrium block in \autoref{eq:zhen_form} strongly enforces that the predicted stress tensor field is both divergence-free (\autoref{eq:cons1}) and symmetric (\autoref{eq:stress_sym}), ensuring the conservation of both linear and angular momentum, respectively.
The equilibrium block is inspired by the physics-encoded Fourier Neural Operator (FNO) introduced by Khorrami \textit{et al.} \cite{khorrami2024divergencefree}, which enforces the divergence-free condition (\autoref{eq:cons1}) but does not ensure the symmetry of the stress tensor.
Mathematically, this block converts an arbitrary vector $P(\boldsymbol{x})$ over the domain $\boldsymbol{x} \in \Omega$ to a symmetric and divergence-free tensor, via the following operations:
}

\begin{equation}
    \boldsymbol{P} (\boldsymbol{x}) = [f(\boldsymbol{x}), g(\boldsymbol{x}), h(\boldsymbol{x})]^T \quad \boldsymbol{x} \in \Omega \subset \mathbb{R}^3
\end{equation}

\begin{equation}
F = \begin{bmatrix}
f_{22} & -f_{12} & 0 \\
-f_{12} & f_{11} & 0 \\
0 & 0 & 0
\end{bmatrix} ,~
G = \begin{bmatrix}
g_{33} & 0 & -g_{13} \\
0 & 0 & 0 \\
-g_{13} & 0 & g_{11}
\end{bmatrix} ,~
H = \begin{bmatrix}
0 & 0 & 0 \\
0 & h_{33} & -h_{23} \\
0 & -h_{23} & h_{22}
\end{bmatrix},
\end{equation}

\begin{equation}
\sigma_{ij} = a F + b G + c H = \begin{bmatrix}
a f_{22} + b g_{33} & -af_{12} & -b g_{13} \\
-a f_{12} & a f_{11}+c h_{33} & -c h_{23} \\
-b g_{13} & -c h_{23} & b g_{11}+c h_{22}
\end{bmatrix}.
\label{eq:zhen_form}
\end{equation}
{
Here, $f_{12} = \partial^2 f / \partial x_1 \partial x_2$. The parameters $a, b, c$ can be selected arbitrarily. Upon inspection of \autoref{eq:zhen_form}, $\boldsymbol{\sigma}$ automatically satisfies the divergence-free requirement and symmetric requirements of Cauchy kinetic variables.
}

{
Both blocks are implemented using finite-difference stencils.
Altogether, these blocks are well suited to solve any solid mechanics problem.
The strongly physics-informed learning problem then reduces to the following simple expression.
}

\begin{equation}
\begin{aligned}
    f_\theta : \boldsymbol{m} \in\mathbb{R}^{R\times N^3} &\mapsto \boldsymbol{\sigma} \in \mathbb{R}^{3\times3\times N^3}, \boldsymbol{\epsilon} \in \mathbb{R}^{3\times3\times N^3} \\
    &\textrm{s.t.} \ \ \ \theta = \arg \min_{\theta} 
    \mathbb{E}_{\boldsymbol{m}, \boldsymbol{\sigma}_{\rm{LR}},\boldsymbol{\epsilon}_{\rm{LR}}} \left[ L_1(D(f_\theta (\boldsymbol{m})), \{ \boldsymbol{\sigma}_{\rm{LR}}, \boldsymbol{\epsilon}_{\rm{LR}} \}) +  L_2(\boldsymbol{C}(\boldsymbol{m}) f_\theta (\boldsymbol{m})[\boldsymbol{\epsilon}], f_\theta (\boldsymbol{m})[\boldsymbol{\sigma}]) \right].
\end{aligned}
\end{equation}
{
In the equation above, $f_\theta(\boldsymbol{m})[\boldsymbol{\epsilon}]$ refers to the $\boldsymbol{\epsilon}$ field predicted by the s-ECO surrogate model, $f_\theta (\cdot)$.
The term $\boldsymbol{C}(\boldsymbol{m})f_\theta(\boldsymbol{m})[\boldsymbol{\epsilon}]$ refers to the transformation of the predicted $\boldsymbol{\epsilon}$ field using the microstructure defined stiffness coefficient field.
The original supervised loss requires only the addition of a soft constitutive-law term executed at the high resolution.
} 

\subsection*{Alternative Implementation}
{
The s-ECO architecture described in the previous subsection automatically ensures mechanical equilibrium and strain compatibility.
Alternatively, it is possible to incorporate the governing physics by weakly satisfying conservation laws; here, the UNet is trained using PINN methods by minimizing a PDE loss (a strategy and network denoted as w-ECO in the Results Section).
It is typical to only penalize $\nabla \cdot \boldsymbol{\sigma}(\boldsymbol{x})=0$ in the loss function.
We found that mechanical compatibility is learned effectively from the supervised training data, so we adopted a similar approach\footnote{We also tried implementations where we filtered the output using the previously described compatibility block. We found no increase in numerical performance but an increase in training instability.}.
}

{
In the w-ECO implementation, a standard UNet directly predicts the kinematic field variable.
The kinetic field is then calculated using the constitutive law.
Specifically, in the first stage, $f_\theta (\boldsymbol{m}) = \boldsymbol{\epsilon}$, and subsequently, $\boldsymbol{\sigma} = \boldsymbol{C}(\boldsymbol{m})f_\theta(\boldsymbol{m})$.
Thus, the overall mapping is expressed as $f_\theta (\boldsymbol{m}) = \{ \boldsymbol{\epsilon}, \boldsymbol{\sigma} \}$.
This simplifies the learning problem into a straightforward expression:
}
\begin{equation}
\begin{aligned}
    f_\theta : \boldsymbol{m}\in\mathbb{R}^{R\times N^3} &\mapsto \boldsymbol{\sigma} \in \mathbb{R}^{3\times3\times N^3}, \boldsymbol{\epsilon} \in \mathbb{R}^{3\times3\times N^3} \\
    &\textrm{s.t.} \ \ \ \theta = \arg \min_{\theta} 
    \mathbb{E}_{\boldsymbol{m}, \boldsymbol{\sigma}_{\rm{LR}},\boldsymbol{\epsilon}_{\rm{LR}}}  \left[ L_1(D(f_\theta (\boldsymbol{m})), \{ \boldsymbol{\sigma}_{\rm{LR}}, \boldsymbol{\epsilon}_{\rm{LR}} \}) + L_2(\Sigma f_\theta (\boldsymbol{m})[\boldsymbol{\sigma}], \boldsymbol{0}) \right].
\end{aligned}
\end{equation}
{
Here, $\Sigma$ is the discrete divergence operator computed using a finite difference stencil.
}

{
While w-ECO weakly enforces the partial differential equations and strongly enforces the constitutive law, s-ECO prioritizes the opposite.
We expect the s-ECO format to offer several practical advantages.
For instance, by swapping the loss functions, the PDE term is removed from the w-ECO loss, which has been shown to result in unstable learning and slow training dynamics \cite{wang2022and}.
}

\subsection*{Architecture}
{
For both case studies, we employed a UNet-based neural surrogate architecture that takes as input a representation of the high-resolution microstructure (i.e. the indicator function map describing the pore or a vectorized stiffness coefficient field, see Supplementary Information (S1) for more details) and predicts the stress and strain fields.
An illustration of the architecture used for the embedded pore case study is shown in \autoref{fig:architecture}.
For the polycrystalline case study, we utilized a larger UNet adapted from the denoising
diffusion probabilistic models (DDPM) architecture \cite{NEURIPS2020_4c5bcfec, buzzy2024statistically}.
In the s-ECO implementation, the UNet output is processed through a compatibility and equilibrium blocks that enforce strain compatibility and stress equilibrium, respectively.
In contrast, the w-ECO implementation uses a 3D convolutional layer to directly map the UNet output to the kinematic tensor field, as shown in \autoref{fig:architecture}.
}

\subsection*{Implementation Specifics}
\label{app:implementationdetails}
{
For the embedded pore case study, the dataset is comprised of tuples $\{ m_n, X_n=\{\sigma_n, \epsilon_n \} \}_{n=1}^{N_{\rm{samples}}}$.
Here, $m_n$ is a high-resolution indicator function field, which takes the value one inside the pore and zero outside the pore.
$X_n$ is a pair of the low-resolution fields.
We aim to train the UNet-based surrogate network, ($\mathcal{G}_{\theta}$), to learn the mapping from $m$ to $X$.
Additionally, since the fields inside the pores are irrelevant, data-driven supervision at low resolution is masked out with a binary mask, $g_n = 1 - m_n$.
The total loss for a given tuple was calculated using following loss function:
}
\begin{align}
    L_{\rm{s\textnormal{-}ECO}}(X_n, \hat{X}_n, g_n) &= \frac{|| g_n (X_n - \hat{X}_n) ||_2^2} {|| g_n X_n ||_2^2} + ||C\epsilon_n-\sigma_n||_2^2 \\
    L_{\rm{w\textnormal{-}ECO}}(X_n, \hat{X}_n, g_n) &= \frac{|| g_n (X_n - \hat{X}_n) ||_2^2}{|| g_n X_n ||_2^2} + ||\Sigma \sigma_n||_1 
\end{align}
{
\noindent where $\hat{X}$ represents the states predicted by s-ECO or w-ECO.
The structure of this loss excludes the contributions from within the pore.
These values are already known analytically, so the network does not need to predict them.
This masking significantly improves the network's training stability.
This is because the network does not have to deal with the abrupt changes at the boundary and can focus solely on accurately estimating the high values within the continuous matrix region. 
}
   
{
The UNet backbone utilized in this case study follows the exact architecture illustrated in \autoref{fig:architecture}.
It comprises five resolution levels, employing 3D convolution and group normalization operations.
Additionally, we used the GELU activation function.
Both the s-ECO and w-ECO models contain 3 million trainable parameters and utilize min-max normalization to normalize the inputs and renormalize the outputs.
The constants $(a,b,c)$ from Eq\@. [12] in the main paper  were assigned the constant values $(1.0, 1.0, 1.0)$.
The networks were trained for 1\,000 epochs using the Adam optimizer \cite{kingma2014adam}, starting with an initial learning rate of $5\times10^{-4}$, which was subsequently decayed using a cosine annealing scheduler.
}

{
For the polycrystalline microstructure case study, the input microstructure, $\boldsymbol{m}_n$, can be either a high-resolution field with dimensions $64^3$ or an ultra-high-resolution field with dimensions $128^3$.
The Euler angles of the microstructure were used to convert a constant cubic coefficient matrix into a coefficient matrix that varies spatially.
This field's $21$ unique coefficients were vectorized.
The variable $X_n= \{\boldsymbol{\sigma},\boldsymbol{F}^e \}$ represents a pair of low-resolution fields, specifically $32^3$, for the Cauchy stress and elastic deformation gradient.
The input coefficient fields underwent a whitening process on a per-channel basis -- i.e. each channel is centered around its mean and normalized by its standard deviation.
These statistics were estimated across the entire training dataset, including all spatial locations.
The output fields were also whitened, but in this case, a single standard deviation was applied across all channels.
This standard deviation was computed by averaging the standard deviations of each channel.
This consistent rescaling ensured that the normalization process did not impact whether a field is compatible or divergence-free. 
Unlike in the other case study with the embedded pore, the loss function did not incorporate masking.
Furthermore, we observed that normalizing the loss function did not affect the model's performance.
}

{
For the s-ECO implementation, the following loss calculation was utilized.
 Here, $\hat{X}_n$ represents the network's prediction \textit{before} downsampling.
 Remember that the s-ECO directly predicts both the kinematic variable, $\boldsymbol{F}^{e}$, and kinetic, $\boldsymbol{\sigma}$.
}

\begin{equation}
\begin{split}
L_{\rm{s\textnormal{-}ECO}}(\hat{X}_n, X_n, \boldsymbol{m}_n) = 
\sum_{i=1}^3\sum_{j=1}^3 \gamma^{1-\delta_{ij}} \Big(
& \alpha \langle \| \sigma_{ij,n,s} - D(\hat{\sigma}_{ij,n,s}) \|_1 \rangle \\
& + \langle \|F^e_{ij,n,s} - D(\hat{F}^e_{ij,n,s})\|_1 \rangle \\
& + \beta \langle \|T(\boldsymbol{C}(\boldsymbol{m}_{n,s}), \hat{\boldsymbol{F}}^e_{n,s})_{ij} - \hat{\sigma}_{ij,n,s} \|_1 \rangle 
\Big)
\end{split}
\end{equation}

{
Here $D(\cdot)$ represents the downsampling operator (we used a windowed average), while $\langle \cdot \rangle$ denotes spatial averaging.
The Kronecker delta $\delta_{ij}$ equals 1 when $i=j$ and is 0 otherwise.
The term $A_{ij,n,s}$ indexes the $ij$ components of the $n^{\mathrm{th}}$ sample of $A$ at voxel $s$.
The hyperparameters $\gamma, \alpha$, and $\beta$ are tunable: $\gamma$ emphasizes the off-diagonal terms, $\alpha$ emphasizes stress, and $\beta$ emphasizes the physics-based loss component (for s-ECO: the constitutive law).
The function $T(\boldsymbol{C}(\boldsymbol{m}), \boldsymbol{F}^e)$ computes the Cauchy stress from the predicted elastic deformation gradient, applied to each voxel.
For voxel $s$, we defined $\boldsymbol{E}_s=0.5(\boldsymbol{F}^{e,T}_s\boldsymbol{F}^e_s-\boldsymbol{I})$ and $T(\boldsymbol{C}(\boldsymbol{m}_s), \boldsymbol{F}^e_s)= |\boldsymbol{F}^{e}_s|^{-1} \boldsymbol{F}^{e}_s\boldsymbol{C}_s \boldsymbol{E}_s \boldsymbol{F}^{e,T}_s$.
Here, $|\cdot|$ is the matrix determinant operation.
Using Einstein notation, this can be expressed as $T(\boldsymbol{C}(\boldsymbol{m}_s), \boldsymbol{F}^e_s)_{ij}= |\boldsymbol{F}^{e}_s|^{-1} F^{e}_{ia,s}C_{abcd,s} E_{cd,s} F_{jb,s}^{e}$.
Since the application involves finite deformations, it is important to ensure that the loss incorporates the correct transformations of the field variables.
}

{
In the implementation of the weak physics-informed U-shaped network, the network only predicts the kinematic variable, $\boldsymbol{F}^{e}$, directly.
The predicted kinetic variable, $\boldsymbol{\sigma}$, is then calculated from this prediction using the method outlined above.
This approach ensures that the constitutive law is strongly enforced.
This w-ECO employed the following loss function.
}


\begin{equation}
\begin{split}
L_{\rm{w\textnormal{-}ECO}}(\hat{X}_n, X_n, \boldsymbol{m}_n) = 
\sum_{i=1}^3\sum_{j=1}^3 \gamma^{1-\delta_{ij}} \Big(
& \alpha \langle \| \sigma_{ij,n,s} - D(\hat{\sigma}_{ij,n,s}) \|_1 \rangle \\
& + \langle \|F^e_{ij,n,s} - D(\hat{F}^e_{ij,n,s}) \|_1 \rangle \\
& + \beta \langle \| (\Sigma \hat{\boldsymbol{\sigma}}_n)_{ij,s} \|_1 \rangle
\Big)
\end{split}
\end{equation}

{
The symbol $\Sigma$ represents the discrete finite difference stencil used to implement the divergence operator.
The term $\hat{\boldsymbol{\sigma}}_n$ refers to the predicted stress at all spatial voxels for the  $n^{\mathrm{th}}$ sample.
}

{
We employed a UNet architecture based on the denoising diffusion probabilistic models (DDPM) \cite{NEURIPS2020_4c5bcfec}, which has proven effective in previous materials science applications\cite{lyu2024microstructure, phan2024generating, buzzy2024statistically}.
We removed the attention mechanism and replaced SiLU activations with ELU activations \cite{rasamoelina2020review}.
To regularize the network, we incorporated dropout with a dropout rate of $0.1513$ before each conditional ResNet block in the DDPM implementation. 
The network starts by expanding the input to $32$ channels and increases the number of channels by factors of $(1, 2, 4, 8)$ after every spatial downsampling.
For the s-ECO implementation, we set the gain of the Kaiming initialization (a method for initializing the weights of neural networks to improve training performance) to $0.5$ to match the marginal distribution of the stress and deformation gradient fields predicted at initialization to the training data.
For s-ECO, we adjusted the gain to $0.2$.
We used the AdamW optimizer \cite{loshchilov2017decoupled} with a weight decay of $0.0165$, excluding learnable parameters in biases, normalization layers, and the `equilibrium' and `compatibility' blocks blocks from this decay \cite{brock2021high}.
AMSGrad \cite{reddi2019convergence} was not utilized.
The learning rate scheduling was implemented with a $100$ epoch linear warm up to a maximum learning rate of $5 \times 10^{-4}$, followed by cosine annealing to a minimum learning rate of $1 \times 10^{-5}$ at $2000$ epochs.
The batch size was set to $16$, and other training hyperparameters were kept at their default values.
The (a, b, c) parameters in the equilibrium block were set as learnable parameters predicted by an MLP from the previous predicted state in the network.
For s-ECO, we used $\alpha = 4.7068$, $\beta=1.3258$, and $\gamma=1.7297$.
For the w-ECO, we used $\alpha = 4.7068$, $\beta=100.0$, and $\gamma=1.7297$.
}

{
The three loss hyperparameters ($\alpha$, $\beta$, and $\gamma$),
the dropout rate, the peak learning rate,
the initial lift channels,
the channel multiplicative factors, and
the weight decay were optimized using the Optuna hyperparameter optimization package \cite{akiba2019optuna}.
We utilized a multivariate tree-structure Parzen estimate sampler \cite{watanabe2023tree} to perform the optimization.
Optuna minimized the loss over a validation dataset of five high-resolution simulations that were permanently excluded from the analysis presented in the main body of this article.
}

\subsection*{Datasets and Problem Descriptions}
{
In order to investigate the practical differences between these two implementations and their implications for physics-based neural super-resolution, we looked at two representative microstructure case studies.
The first case study looks into the prediction of the mechanical fields surrounding an arbitrarily shaped pore embedded in a continuum domain.
For this problem, the low and high resolutions were a $128^3$ and a $256^3$ voxel mesh, respectively.
This dataset consists of 2\,500 infinitesimal-strain fast Fourier transform (FFT) simulations \cite{lebensohn2012elasto} under periodic boundary conditions using a 80:10:10 training, validation and testing split.
The training and validation datasets' simulated mechanical fields (i.e. the stress and strain fields) are available in the lower resolution (LR), while in the testing dataset, higher-resolution simulations are performed in order to test the trained surrogate's performance.
Additional information on this dataset is provided below.
}

{
The second case study consists of polycrystalline microstructures deformed elastically.
For this problem, we considered three resolutions: $32^3$, $64^3$, and $128^3$ -- low, high and ultra-high resolutions, respectively.
This dataset is composed of 1\,000 crystal plasticity simulations \cite{alleman2018} at the low resolution.
Additionally, we performed $20$ and $10$ simulations at the high and ultra-high resolutions for testing.
The low-resolution simulations were split 75:25 into training and validation data.
Additional information on this dataset is provided below.
}

\subsection*{Fast Fourier Transform (FFT) Simulations for Embedded Pores Case Study}
\label{app:spectralsim_methodology}
{
For the the embedded pore case study, we employed a parallelized version of the elasto-viscoplastic fast Fourier transform (FFT) simulation, known as Micromechanical Analysis of Stress-Strain Inhomogeneities with Fourier Transform (MASSIF), as described in \cite{lebensohn2012elasto}, to evaluate the stress and strain response of individual pores under uniaxial tension.
The pores were extracted from micro X-ray computed tomography ($\mu$XCT) scans of additively manufactured Ti-6Al-4V parts produced via laser powder bed fusion. 
This dataset was created as part of a NASA University Leadership Initiative (ULI) program in collaboration with Carnegie Mellon University and includes $23$ $\mu$XCT scans.
Individual pores were grouped based on visual similarity, and stratified sampling by cluster identity was used to create a balanced dataset of 2\,500 pores, ensuring representation of all characteristic pore shapes.
The dataset was split into a training set of 2\,000 pores, a validation set of 250 pores, and a test set of 250 pores.
Training and validation simulations were conducted at a resolution of $128^3$, while test simulations were performed at both $128^3$ and $256^3$.
The $128^3$ pores were downsampled to a $64^3$ low resolution, and the $256^3$ pores were downsampled to a $128^3$ high resolution.
The MASSIF simulations applied a single step of elastic deformation under Z-tension to a total strain of $1 \times 10^{-3}$, with a maximum of 100 iterations and a convergence criterion of $1 \times 10^{-6}$.
}

\subsection*{Crystal Plasticity Finite Element Simulations for Polycrystalline Microstructures Case Study}
\label{app:femsim_methodology}
{
For the the polycrystalline microstructure case study, we generated 1\,000 unit-volume microstructures using the DREAM.3D software package \cite{groeber2014}, targeting a lognormal grain size distribution with parameters $\mu = -0.788$, $\sigma = 0.0998$.
The number of grains in each microstructure varied between $21$ and $28$, with an ensemble average of $\sim 24$.
The bulk of the microstructures were generated on a cubic grid at the low resolution, $32^3$.
The ultra-high resolution microstructures were generated at a native resolution of $128^3$ and downsampled to the high ($64^3$) and low resolutions by assigning grain ids to voxels in lower-resolution models according to the maximum volume fraction in the corresponding higher resolution voxels.
We performed finite element analysis for each microstructure to simulate elastic deformation loaded in the tensile $x$-direction up to a strain of $8 \times 10^{-5}$.
The elastic material model used was that of Alleman and coworkers\cite{alleman2018}, which represents the properties of austenitic stainless steel.
The linear elastic constitutive relationship employs cubic symmetry to evoke the performance of the face-centered cubic crystal, with elastic constants (in GPa) $C_{11} = 204.6$, $C_{12} = 137.7$, $C_{44} = 126.2$.
}

\newpage
\section*{Acknowledgements}
V. Oommen expresses gratitude to Prof\@.~W. Curtin at Brown University for his insightful suggestion on masking out the embedded pores and for his many other valuable discussions. The authors acknowledges D. Vizoso from Sandia National Laboratories for his valuable insights and discussions on the work and manuscript.

\noindent \textbf{Funding:} 
D. Diaz and A.D. Rollett are grateful for support from the NASA STRI program under Award Number 80NSSC23K1342. They also acknowledge the longstanding collaboration on spectral methods for micro-mechanics with Dr. Ricardo Lebensohn (LANL).
This work is supported by the Center for Integrated Nanotechnologies, an Office of Science User Facility operated for the U.S. Department of Energy. This article has been authored by an employee of National Technology \& Engineering Solutions of Sandia, LLC under Contract No. DE-NA0003525 with the U.S. Department of Energy (DOE). The employee owns all right, title, and interest in and to the article and is solely responsible for its contents. The United States Government retains and the publisher, by accepting the article for publication, acknowledges that the United States Government retains a non-exclusive, paid-up, irrevocable, world-wide license to publish or reproduce the published form of this article or allow others to do so, for United States Government purposes. The DOE will provide public access to these results of federally sponsored research in accordance with the DOE Public Access Plan https://www.energy.gov/downloads/doepublic-access-plan.

\noindent \textbf{Author Contributions:} 
V.O., A.R., G.K., and R.D. conceived the idea and designed the research methods.
V.O. and A.R. implemented the machine-learning model.
Z.Z. developed concept for machine-learning architectural block.
D.D., A.D., C.A. generated the training data.
V.O. and A.R. analyzed the results and validated the results.
R.D. and G.K. supervised and guided the research project.
R.D., A.D., and G.K. secured funding for the research project.
V.O., A.R., and R.D. prepared the initial draft.
All authors reviewed and revised the manuscript.

\noindent \textbf{Competing Interests:} 
The authors declare no competing interests. 

\noindent \textbf{Data and Materials Availability:} 
The codes will be made available after acceptance at our \href{https://github.com/vivekoommen/PhysicsBased_NeuralSR}{GitHub Repository}\cite{ECO2025}

\newpage
\bibliographystyle{unsrt}  
\bibliography{references}  
\newpage

\appendix
\counterwithin{figure}{section}
\setcounter{figure}{0}


\section{Representations of Polycrystalline Microstructures for Localization Problems}
\label{app:polycrystalline_inputs}
{
A good representation of the input is essential for a successful deep learning model.
In this study, we aimed to establish a direct mapping between microstructure and the resulting stress and strain fields under constant boundary conditions.
This mapping is well understood in single-resolution settings for ``N-phase'' problems, where the microstructure consists of several constant phases.
The embedded pore case study is an example of a 2-phase problem, but there are fewer studies on polycrystalline systems.
Notably, existing examples often require large datasets; for instance, \textit{et al.} \cite{frankel2020prediction} needed $16,000$ input-output pairs to train a neural surrogate for polycrystalline materials, while Saha \textit{et al.} \cite{brady2024unet} used only $420$ samples to train a U-Net for a two-phase problem.
}

{
During our experimentation with polycrystalline data, we found that the complexity of the target map varied significantly depending on the chosen representation of the microstructure.
The best results were achieved by using stiffness coefficients as direct inputs to the microstructure.
Specifically, we computed a dense stiffness tensor field from the Euler angle field and the cubic stiffness tensor of the material using standard tensor transformations.
We then vectorized the $21$ unique coefficients of this tensor field to serve as input to the neural network, \textit{de facto} effectively featurizing the Euler angles.
This approach was inspired by the work of Khorrami \textit{et al.} \cite{khorrami2023unet, khorrami2024divergencefree}, who studied polycrystal-like microstructures where each grain was represented by an isotropic elastic stiffness tensor.
They varied the tensor parameters, such as Young's modulus and Poisson's ratio, between grains to simulate real polycrystalline materials.
By using these parameters as direct inputs, they achieved data scaling similar to N-phase problems.
Surprisingly, we found that using the vectorized stiffness coefficients as inputs worked exceptionally well, even though it dramatically increased the input's dimensionality.
A similar concept was explored by Kelly and Kalidindi \cite{kelly2024therino}.
}

{
Here, we briefly compare different representation schemes for polycrystalline microstructures.
To streamline our analysis, we focus on learning the mapping between the microstructure and the stress component $\sigma_{11}$ only.
We limit our study to a low-resolution learning problem, defined as $f:\mathbb{R}^{R\times32\times32\times32} \mapsto \mathbb{R}^{1\times32\times32\times32}$, where $R$ represents the dimension of the input representation.
Additionally, we used a purely data-driven training approach, without incorporating any physics-informed elements (weak or strong form).
We compare three input representations:
(1) the vectorized stiffness coefficient field ($R=21$), 
(2) Euler angles ($R=3$), and
(3) reduced order generalized spherical harmonics (ROGSH) ($R=3$) \cite{buzzy2024statistically}.
 It is crucial to compute the ROGSH coefficients from the Euler angles using a consistent convention; for instance, Buzzy \textit{et al.} \cite{buzzy2024statistically} derived ROGSH coefficients based on the standard intrinsic ZXZ convention.
 We performed hyperparameter tuning for each input representation but found that performance remained largely unaffected by hyperparameter choices.
 To prevent overfitting, we applied a high dropout rate ($r=0.4$) for the Euler angle and ROGSH inputs.
} 

\begin{figure*}[htbp!]
  \centering
  \includegraphics[width=1.0\textwidth]{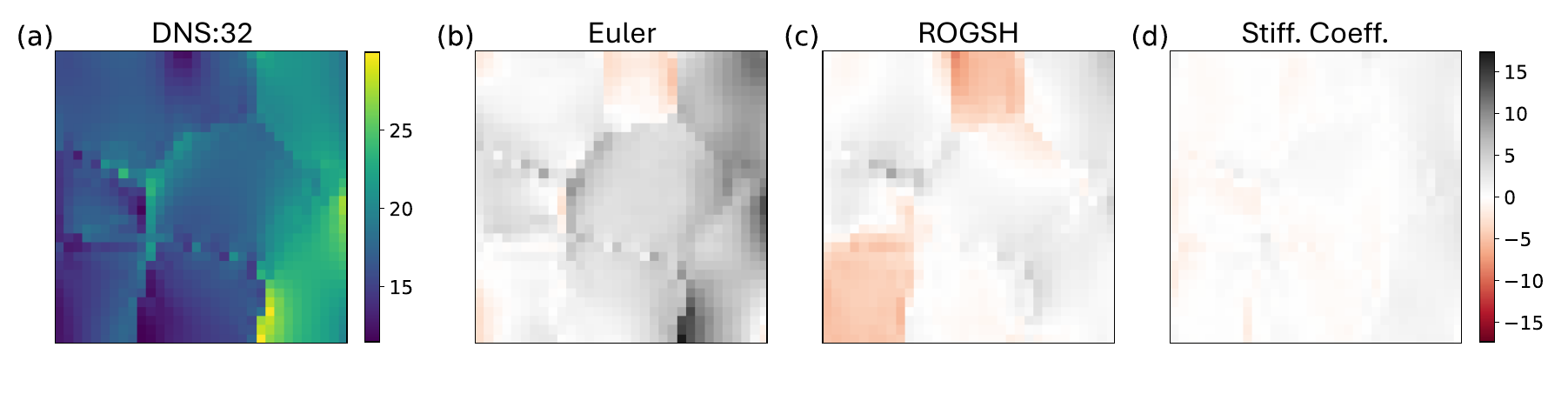}
  \caption{
  2D Slice of $\sigma_{11}$ field.
  Comparison between ground-truth, low-resolution ($32\times32\times32$) simulation and error in predictions from three machine-learning models with three different microstructure-representation inputs --
  (a) Euler angles,
  (b) reduced order generalized spherical harmonics (ROGSH), and
  (c) vectorized stiffness coefficients.}
  \label{fig:appendix_A_comparison}
\end{figure*}
\begin{figure*}[htbp!]
  \centering
  \includegraphics[width=0.5\textwidth]{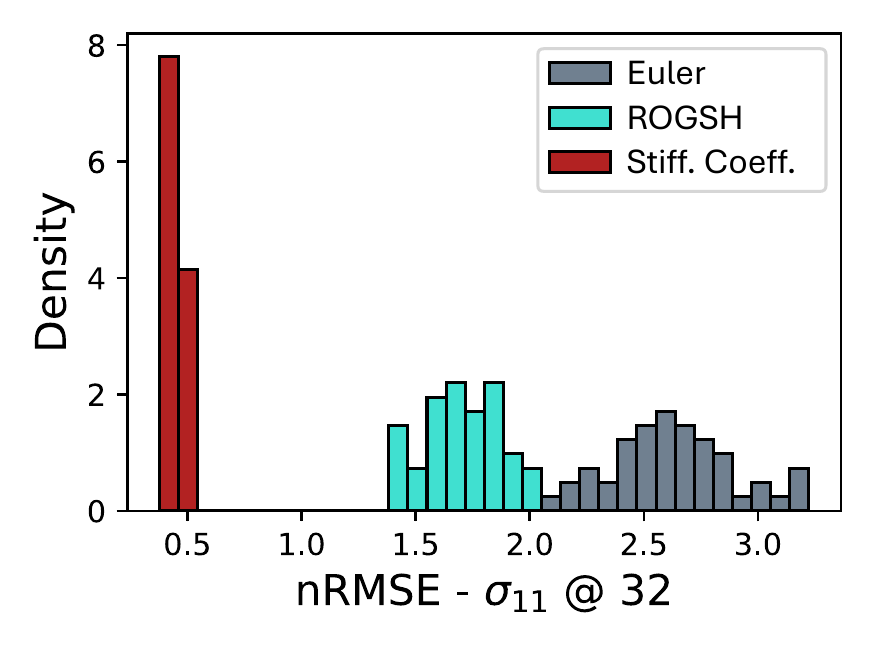}
  \caption{
  Model performance over testing dataset of $\sigma_{11}$ fields at low resolution of networks trained with stiffness coefficient, Euler angles, and reduced order generalized spherical harmonics (ROGSH) as inputs.
  nRMSE is the root mean squared error computed on the normalized predictions.}
\label{fig:appendix_A_dataset_distribution}
\end{figure*}

{
\autoref{fig:appendix_A_comparison} contrasts rom models trained with the three input representations.
The model using vectorized stiffness coefficients accurately predicts the $\sigma_{11}$ field, showing mostly uniform and low errors.
After extensive fine-tuning, the model using the ROGSH representation captures some dominant features but struggles to resolve grain boundaries and fails to predict off-diagonal components.
In contrast, the model using the Euler-angle representation does not provide meaningful predictions.
This performance trend is consistent across the entire dataset, as shown in \autoref{fig:appendix_A_dataset_distribution}.
Clearly, the stiffness coefficient parameterization significantly outperforms the other approaches.
}

{
We hypothesize that the observed improved performance occurs because the transformation from Euler angles to stiffness coefficients is of high polynomial order:
}

\begin{equation}
    C(x) = g(\Phi(x)) = g((\phi_1, \Psi, \phi_2)(x)) = RRCR^TR^T
\end{equation}

{
Here, $C$ represents the constant stiffness matrix of the material system, and $R$ is the rotation matrix.
The rotation matrix $R$ is the product of three individual rotation matrices, which exact form depends on the selected Euler angle convention: $\hat{R}(\phi_1)\hat{R}(\Psi)\hat{R}(\phi_2)$.
The transformation can involve up to the $4^{\mathrm{th}}$ powers of trigonometric functions of the individual Euler angles and can include up to $12$-term products.
By using the (rotated) stiffness coefficients as inputs, we avoid this complexity and simplify the learnable mapping.
}

\section{Extended Analysis: Embedded Pores}
\label{app:extendedanalysis}

\begin{figure*}[htbp!]
  \centering
  \includegraphics[width=1.0\textwidth]{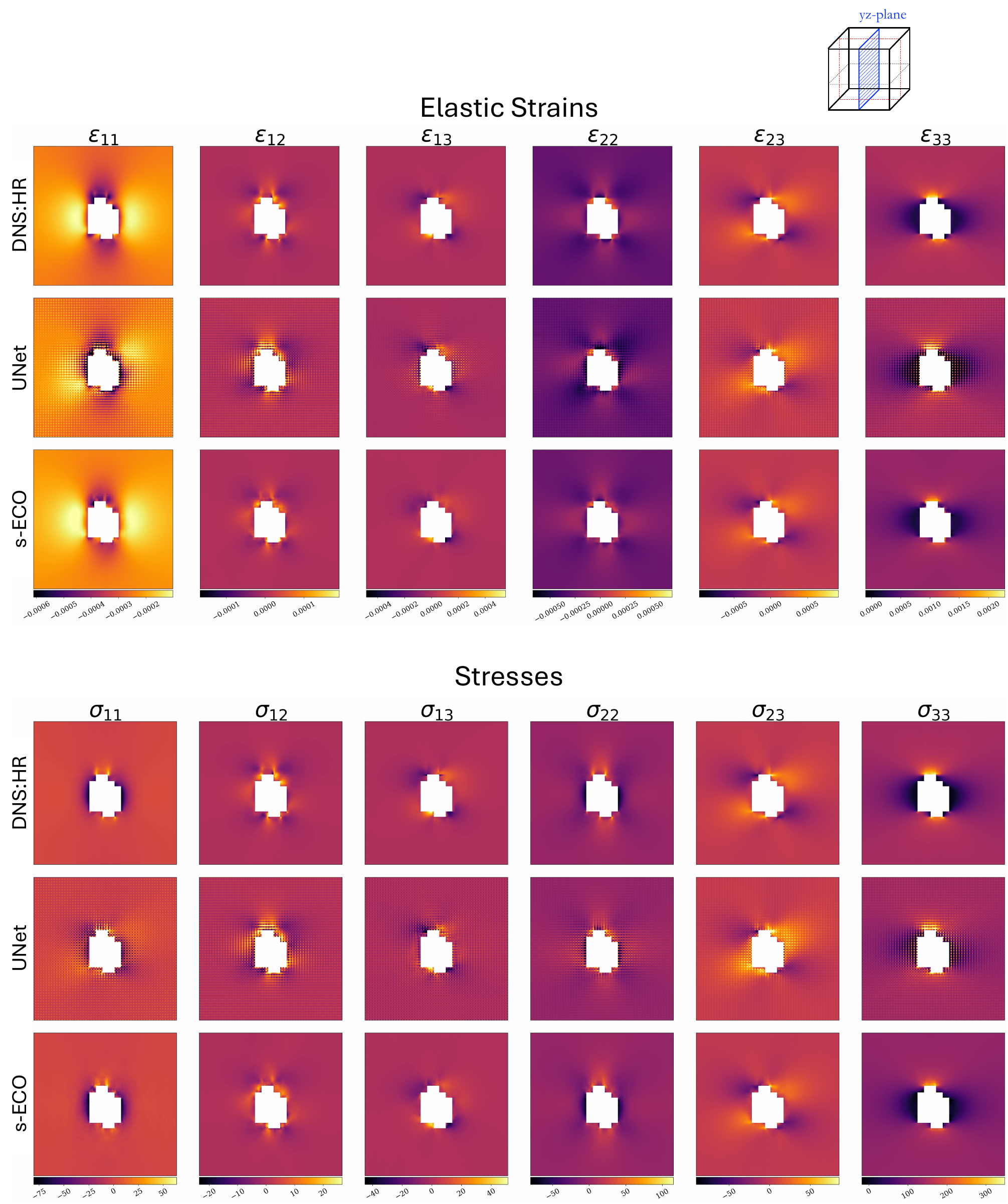}
  \caption{
  Comprehensive comparison of center slice from all tensor components between the high resolution direct numerical simulation (DNS:HR) and predictions from networks trained in the super-resolution format: s-ECO (using proposed physics-based neural super-resolution framework) and UNet (using super-resolution format without physics-based components). Without the physics based addition, the UNet predictions display clear pathologies (e.g., checkerboard artifacting).}
  \label{fig:spunetvsunet}
\end{figure*}

In the main manuscript, we analyzed the importance of incorporating physics in super-resolution training. We focus on the dominant component for our problems -- the $xx$-direction components, however, the findings are consistent across the remaining values. In \autoref{fig:spunetvsunet}, we see that across all components, the UNet implementation -- trained in the super-resolution format \textit{without} physics-based supervision -- displays persistent checkerboarding artifacts. We found that the structure of these artifacts depends on the utilized downsampling operation, but that their presence is persistent across all options tried. In contrast, the physics-informed super-resolution (represented here by s-ECO) successfully predicts across all components. We refer the reader to Table 1 in the paper's main body for summary statistics across the entire testing dataset. 

\section{Mixed-Precision Acceleration}
\label{app:mixed_precision}

\begin{figure*}[htbp!]
  \centering
  \includegraphics[width=1.0\textwidth]{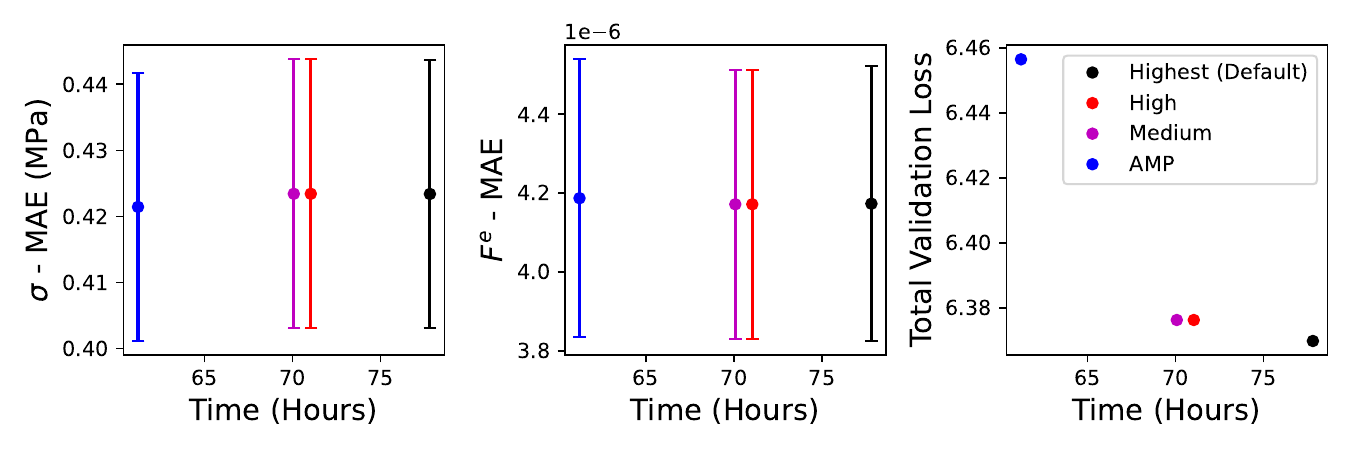}
  \caption{
  Final inference performance on high-resolution test dataset ($64^3$) for the s-ECO implementation applied to the polycrystalline microstructures case study.
  Comparison between the standard implementation and equivalent architectures trained using three variants of mixed precision training.
  MAE stands for mean absolute error.}
  \label{fig:mixedprecision}
\end{figure*}

{
Neural surrogates for solving parametric PDEs are are amongst the more complex deep learning problems encountered in Materials Informatics.
They require
large architectures,
extensive training datasets, and
high-resolution inference,
which can significantly increase computational costs and training times.
 In this study, we employed several strategies to reduce training time, with multi-processing data loading being the most effective.
 This simple option in PyTorch allowed us to distribute the computational load of loading and preprocessing batches across multiple CPU cores.
 We also used mixed precision training to reduce the computational cost of forward inference and back-propagation.
 Since our training environment was primarily limited by CPU speed (i.e., the rate at which we could load individual batches), we saw only limited acceleration from mixed precision training.
 However, in scenarios where data can be loaded quickly, we anticipate a significant computational benefit.
}

{
In most cases, we found that the proposed strong physics-informed U-shaped network (s-ECO) architecture displayed stable performance with mixed precision.
In this subsection, we provide a brief benchmark of mixed precision training for the s-ECO implementation.
We compared PyTorch's Automatic Mixed Precision (AMP) framework, adjusting multiplication precision ($\mathrm{set\_float32\_matmul\_precision}$ (``medium''$|$``high'')), and the standard training ($\mathrm{set\_float32\_matmul\_precision}$ (``highest'')).

For the polycrystalline microstructure case study (from low resolution to high resolution, $32 \rightarrow 64$), we observed almost no change in final performance across the four implementations, as shown in \autoref{fig:mixedprecision}.
Even in our memory-limited computational environment, all variants reduced training time compared to the baseline.
However, in the embedded pore case study, AMP training introduced significant checkerboarding artifacts.
We hypothesize that these artifacts were due to gradients in the fields exceeding the precision limit of $16$-bit numbers.
The other acceleration methods yielded stable results for both case studies.
}

\end{document}